# Detail Loss in Super-Resolution Models Based on the Laplacian Pyramid and Repeated Upscaling and Downscaling Process [†]

## Sangjun Han[1] | Youngmi Hur[2]

[1]School of Mathematics & Computing (Mathematics), Yonsei University, Seoul, South Korea

[2]Department of Mathematics, Yonsei University, Seoul, South Korea

**Correspondence**

Youngmi Hur, Department of Mathematics, Yonsei University, 50, Yonsei-ro, Seoul, South Korea.
Email: yhur@yonsei.ac.kr

## Abstract

With advances in artificial intelligence, image processing has gained significant interest. Image super-resolution is a vital technology closely related to real-world applications, as it enhances the quality of existing images. Since enhancing fine details is crucial for the super-resolution task, pixels that contribute to high-frequency information should be emphasized. This paper proposes two methods to enhance high-frequency details in super-resolution images: a Laplacian pyramid-based detail loss and a repeated upscaling and downscaling process. Total loss with our detail loss guides a model by separately generating and controlling super-resolution and detail images. This approach allows the model to focus more effectively on high-frequency components, resulting in improved super-resolution images. Additionally, repeated upscaling and downscaling amplify the effectiveness of the detail loss by extracting diverse information from multiple low-resolution features. We conduct two types of experiments. First, we design a CNN-based model incorporating our methods. This model achieves state-of-the-art results, surpassing all currently available CNN-based and even some attention-based models. Second, we apply our methods to existing attention-based models on a small scale. In all our experiments, attention-based models adding our detail loss show improvements compared to the originals. These results demonstrate our approaches effectively enhance super-resolution images across different model structures.

## 1 | INTRODUCTION

As hardware has advanced in recent years, image-processing techniques for handling high-resolution (HR) images have become increasingly indispensable. One such technique is the single image super-resolution (SR), an image enhancement method that generates an HR image from a low-resolution (LR) one. It has attracted considerable interest because of its wide-ranging applications, including medical imaging [1–3], object detection [4, 5], and satellite image analysis [6, 7].

Deep-learning methods, which have become a central focus of recent research, are widely applied to image processing. When integrated into super-resolution frameworks [8–12], these methods yield marked improvements in performance.

Researchers have explored various approaches, such as developing deeper convolutional neural networks (CNNs) [13–15] and designing algorithms that incorporate existing image-processing techniques [16–20]. Recently, approaches using attention-based structures [21–26] have also been proposed.

Many recent deep-learning approaches concentrate on extracting informative features from an LR image and effectively propagating this information throughout the entire SR pipeline. For instance, the networks introduced in [19, 22, 26] employ customized attention modules–such as holistic and Laplacian attention–to enhance their feature extraction capabilities. Other studies [27, 28] extend the conventional mean squared error or $l_1$ loss between SR and HR images by incorporating auxiliary objectives explicitly designed to encourage the recovery of fine details. Inspired by the demonstrated importance of capturing high-frequency information, we also introduce an additional auxiliary loss. Our ultimate goal is to enhance high-frequency fidelity in SR while keeping the framework simple and broadly applicable, and additional auxiliary loss is a key step toward that objective.

Specifically, we propose a detail control loss based on the Laplacian pyramid (LP) to guide the high-frequency part of SR. Our method leverages the reconstruction concept of the LP, which returns the high-resolution source image by adding an upsampled approximation image with a detail image (referred to as the error image in the paper [29]). Our method creates a feature map for the detail image from the upsampled features and controls it separately from the SR features by introducing an additional loss function. This approach allows the model to activate meaningful pixels for high-frequency details and focus more on generating these fine details. Additionally, we apply a repeated upscaling and downscaling process (RUDP). Our RUDP repeats downsampling the completed SR feature map and then combining it with the LR image to extract new SR and detail features. Our experiments demonstrate that combining RUDP with the LP-based detail loss effectively extracts various information from the LR image.

We conduct two main experiments. First, we construct a simple CNN-based model, Laplacian pyramid-based Upscaling and Downscaling super-resolution network (LaUD), that incorporates our two methods. This CNN-based model outperforms all currently available state-of-the-art (SOTA) CNN models and even outperforms several attention-based approaches when evaluated in terms of PSNR and SSIM. Moreover, our ablation study and feature map analyses demonstrate that our detail loss and RUDP are effective methods for improving performance. We also confirm that their effectiveness is further enhanced when both methods are used together. Additionally, we assess LaUD with perceptual quality metrics beyond PSNR and SSIM, confirming its perceptual fidelity. Second, we apply our methods to existing attention-based models on a small scale. Comparing the results with and without them, we observe that their use consistently enhances performance across all models. These results show that our methods are applicable both with and without attention mechanisms and can also improve the performance of attention-based models even further.

In summary, our main contributions are the following:

- We propose a new method, the detail control loss based on the Laplacian pyramid. This method allows the model to handle the detail image for high-frequency information apart from the SR image. Consequently, the model can focus more on the detail part and supplement information that is not present in the upsampled image.
- We also introduce a new method, which we call RUDP, and show that RUDP effectively integrates with our LP-based detail loss. The analysis of the SR feature maps demonstrates that RUDP allows the model to capture more diverse information by re-extracting features from the SR features supplemented with details.

- We apply our two methods to both CNN-based and attention-based models. As a result, we establish their effectiveness across all models, regardless of structure.

Although our two contributions, the LP-based detail control loss and RUDP, can be integrated as modules within existing SR models, we will refer to them as methods, since they are model-independent and together encompass both loss-based training strategy and architectural components.

## 2 | RELATED WORK

### 2.1 | From CNNs to Attention

In the early days of deep learning for SR, many studies [8, 9, 13, 30] aimed to deepen models more efficiently. The paper [13] is a pioneer in this direction, introducing deeper structures through residual learning. Subsequently, several papers have developed efficient models based on residual networks. Authors of [14] build the model EDSR to improve performance by constructing a multi-scale structure with residual blocks, and the work [15] proposes a cascade connection between residual blocks to efficiently produce SR images with fewer parameters. Similar to these studies, our CNN-based model, LaUD, utilizes residual blocks and skip connections to deliver information efficiently. Moreover, our RUDP method enables LaUD to extract more diverse features within a deep architecture for better SR performance.

Recently, the transformer model has demonstrated excellent feature extraction performance and has been successfully adapted for visual tasks [31–34]. Consequently, many studies utilize the attention mechanism in the SR task. Authors of [22, 35] enhance performance by using both channel-wise and spatial-wise attention. The DRLN model [19] employs channel attention with a pyramid concept to capture different sub-frequency-band information. The paper [23] presents EDT, a model obtained by modifying the window shape to enhance connectivity between windows. Some papers, such as [25, 36], utilize transformer models that have demonstrated high performance in the visual domain. For our methods, experiments in which they are applied to existing attention-based models illustrate both the applicability of our approach and the benefits of combining it with attention mechanisms.

### 2.2 | Loss Functions for the SR Task

Fine details, often absent in LR images, pose a significant challenge in SR problems. To overcome this, many studies have sought to enhance performance by introducing various loss functions beyond traditional ones, such as mean squared error (MSE) between SR and HR images. In [37], the model generates multiple SR images and aggregates their MSE losses. Some papers, such as [28, 38], introduce an additional loss based on the feature maps of a pretrained model. Since a well-trained model captures the style of an image, including texture and patterns, its feature maps can help mitigate deficiencies in the high-frequency parts. While their incorporation of additional loss functions beyond the standard difference between SR and HR images is similar to our approach, our method employs an LP-based detail image to guide the model in capturing high-frequency information. In [27, 39], high-frequency components are supplemented by leveraging frequency-domain information. Specifically, the paper [27] applies a Fourier transform to the HR image for frequency-domain supervision. However, such frequency-domain information generally lacks, or fails to preserve, spatial information, potentially causing inaccuracies or misplacement of high-frequency details. In contrast, our LP-based detail loss simultaneously provides high-frequency and spatial information without the global application of frequency-component enhancement, effectively preventing the aforementioned issues. In [40, 41], the authors extract detailed parts of images for new loss functions through edge detection and gradient extraction convolution. In particular, although the approach in [40] is somewhat similar to ours in extracting edge images from HR images, it differs from our detail control as its edge images are not involved in the reconstruction process of SR images.

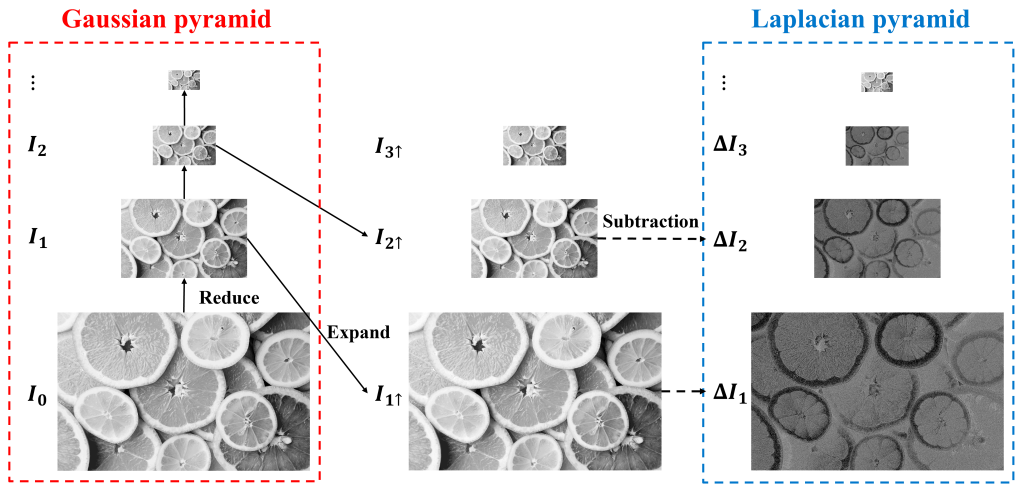

**FIGURE 1** The construction process of the Laplacian pyramid. (Source: freestocks/Unsplash, Unsplash License)

## 2.3 | LP Structure in the SR Problem

There have been many attempts to combine mathematical theories with deep learning. Given that wavelets can handle multi-resolution images and integrate naturally with a convolution layer, various research [17, 42, 43] have been conducted. They generate high-frequency images of the same size from the LR input and apply the inverse wavelet transform to produce an SR image. In contrast, we employ the LP reconstruction process. The LP detail image, which is the same size as the HR image, contains more information. Combined with RUDP, this leads to enhanced abundance and diversity in feature extraction. In [16, 19, 44], the authors introduce the pyramid structure of LP in their models. Specifically, the Lap-SRN model [16] utilizes a pyramidal reconstruction structure in LP. Although the strategies for generating details and the reconstruction process are similar to ours, our approach differs from LapSRN in that we incorporate the detail into a loss function, guiding the model to better capture high-frequency data. In DRLN [19], the authors propose the Laplacian attention that generates feature maps of different scales similar to the pyramid structure of LP and uses them as channel attention. Unlike DRLN, we directly control the detail feature map through the loss function and consider the LP structure solely in the reconstruction process.

## 3 | METHODS

### 3.1 | Laplacian Pyramid

The Laplacian pyramid [29] is an image representation consisting of multi-scale high-frequency images and one low-frequency image of the smallest scale. This representation is similar to the Gaussian pyramid presented in the same paper, but differs in that the Laplacian pyramid comprises residual images except for the last level.

Figure 1 illustrates the overall process for constructing the Laplacian pyramid. First, we obtain a downscaled image $I_1$ by low-pass filtering and downsampling an original image $I_0$. Then, $I_1$ is expanded to the same resolution as $I_0$ via interpolation, yielding a re-upscaled image $I_{1\uparrow}$. Subtracting $I_{1\uparrow}$ from $I_0$ produces a residual image $\Delta I_1$, which we refer to as the "detail image." Consequently, the original image is decomposed into the approximation image $I_1$ and the detail image $\Delta I_1$, forming the first level of the LP. Repeating this process to the approximation image $I_n, n = 0, 1, \ldots, k - 1$, we create a pyramid composed of $k$ multi-scale high-frequency images $\Delta I_1, \Delta I_2, \ldots, \Delta I_k$ and one low-frequency image $I_k$ of the smallest size, after $k$ steps. Since the process of constructing the Laplacian pyramid involves subtraction, the LP can completely reconstruct the high-resolution source image by adding a detail image and an upsampled approximation image of the same level. Therefore, LP is a useful technique for image compression and reconstruction.

We consider the residual image of the LP suitable for refining the high-frequency components of the SR image. If a model generates an elaborate LP detail image, the perfect reconstruction property of LP is exploited effectively to enhance the SR output. From this perspective, we develop an LP-based detail control approach. This approach guides a model to generate a feature map enriched with high-frequency information, using supervised learning on the LP detail of the HR image.

## 3.2 | LP-based Detail Loss

Numerous models proposed for the SR task have explored various approaches aimed at enhancing high-frequency information. Among these methods, modifying the loss function represents an efficient and intuitive strategy that can be seamlessly integrated into diverse model architectures. It is well-known that relying solely on conventional losses such as $l_1$ or MSE between HR and SR images often results in overly smoothed outputs, yielding lower-quality SR images. To mitigate this issue, we introduce an additional loss specifically designed to control image details, thereby encouraging the model to better capture and emphasize high-frequency components.

The LP introduced in Section 3.1 is a useful image processing tool that excels in image analysis and enables perfect reconstruction. Although wavelet and Fourier transforms are also powerful image processing techniques, LP provides several comparative advantages. Specifically, it is straightforward to implement, requiring only simple low-pass filtering followed by a differencing operation. Additionally, high-frequency information at each LP level can be clearly inspected without frequency scrambling, and direct manipulation of pixel values at individual pyramid levels leads naturally to intuitive and predictable reconstructions. For additional discussion of these properties, see [45] and references therein. Motivated by these strengths, we propose training a model with an additional loss function that utilizes the LP-based detail image as ground-truth, thereby encouraging the model to more effectively capture high-frequency components.

Among the detail images produced at each level of the LP, we utilize only the largest one (mainly for simplicity), which matches the size of the HR image. Specifically, our LP-based detail loss is defined as the $l_1$ difference between the detail image $D_{SR}$ produced by the model and the ground-truth LP detail image $D_{GT}$. Thus, our total loss $L$ is constructed by augmenting the conventional $l_1$ loss between the HR image $I_{HR}$ and the SR image $I_{SR}$ with this LP-based detail loss, as follows:

$$L = \|I_{HR} - I_{SR}\| + \lambda \cdot \|D_{GT} - D_{SR}\|. \qquad (1)$$

Here, $\lambda$ denotes a hyperparameter that determines the balance between the conventional loss and the LP-based detail loss. Typically, methods employing auxiliary losses heavily weight the conventional loss to preserve PSNR values, leading to limited high-frequency enhancement. In contrast, our approach maintains robust PSNR values even at higher $\lambda$. Therefore, we set $\lambda = 1$ for all experiments conducted in this paper to maximize the model's emphasis on high-frequency information.

To train the model with our proposed loss, the model must generate a detail image. To achieve this, we introduce one type of upscale block structure within our CNN-based model, LaUD (cf. Section 3.4.1). In particular, we construct the upscale block (cf. Figure 2) to closely resemble the reconstruction process of LP. We anticipate that this design enables the model to effectively incorporate high-frequency information into the SR image by directly combining the model-generated detail feature map with the upscaled feature map. Furthermore, this upscale block structure can also be easily incorporated into attention-based models constructed from scratch, enabling the utilization of the LP-based detail loss by leveraging the inherent reconstruction properties of LP.

On the other hand, for existing models, incorporating the above upscale block may be challenging due to structural constraints of the original architecture. Nevertheless, it remains feasible for such models to generate a detail image. Specifically, an auxiliary upsampler that produces the detail image from extracted features can simply be appended to the final image-generation stage of the existing model. In our experiments involving existing attention-based models, where direct integration of the upscale block designed in LaUD was impractical, we successfully employed this alternative strategy and achieved promising results.

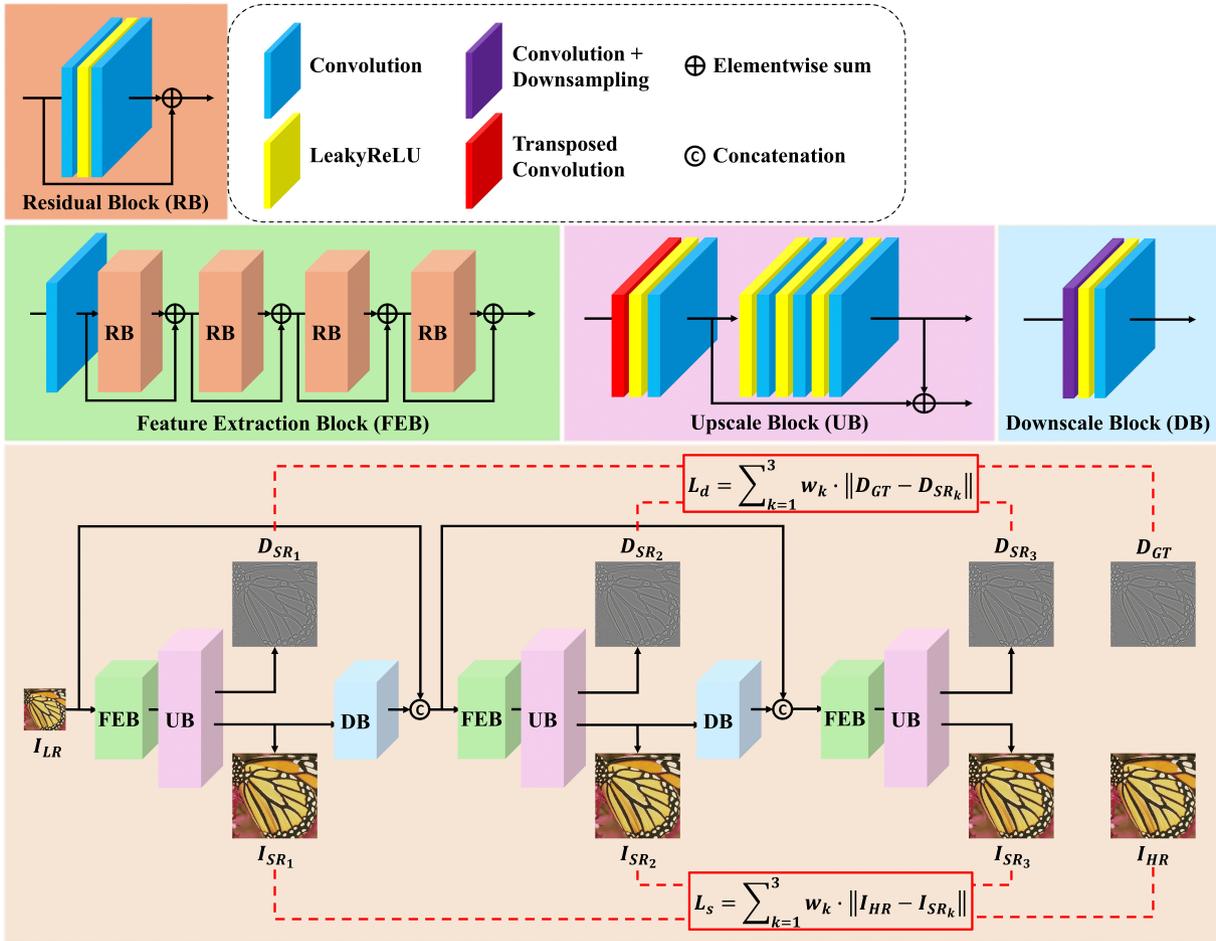

**FIGURE 2** Structure of our CNN-based model, LaUD. In the main LaUD structure (bottom), $D_{SR_k}$ denotes the generated detail images, and $D_{GT}$ denotes the ground-truth LP detail image. Similarly, $I_{SR_k}$ are the generated SR images and $I_{HR}$ is the ground-truth HR image. $L_d$ represents our LP-based detail loss, and $L_s$ indicates the traditional image loss computed between HR and SR images.

## 3.3 | Repeated Upscaling and Downscaling Process

We propose the Repeated Upscaling and Downscaling Process (RUDP) as a structural approach to further enhance the effectiveness of the LP-based detail loss. Specifically, once the model generates an SR feature map, RUDP produces a new low-resolution feature map by downsampling the SR feature map. The model then iteratively repeats the process of extracting features from this downsampled representation and reconstructing an updated SR image. The number of repetitions of the up-and-down process is treated as a hyperparameter, and the model's final SR image is obtained after completing these repetitions.

In RUDP, the concatenation of the original LR image and the newly obtained low-resolution feature maps serves as the input for feature extraction in the subsequent step. By using multiple low-resolution inputs, we expect the model to leverage a broader range of information, which can help produce a more refined SR image. Furthermore, we utilize all intermediate SR images generated during RUDP to construct the total loss function, allowing the model to produce reliable SR images at each intermediate step. This strategy provides intermediate supervision, potentially making it easier for deeper models to produce a reasonable final SR image. In addition, the model can refer to SR images obtained from previous RUDP steps, identifying and compensating for any deficiencies in these intermediate outputs to reduce training loss. As a result, this approach supports the progressive refinements of SR images throughout RUDP.

When used with our LP-based detail loss, RUDP can yield a beneficial effect on model performance. According to our analysis, the detail loss helps the model focus on high-frequency information, thereby facilitating the extraction of meaningful features for the SR image. Meanwhile, RUDP strengthens these extracted features by concatenating them back into the input after downsampling. Hence, integrating these two methods can lead to improved performance, as supported by our experimental results.

In our LaUD model, the structure used for RUDP comprises an upscale block and a simple downscale block, as illustrated in Figure 2. We configure LaUD to execute the upscaling process three times. Since LaUD also produces detail images required for our LP-based detail loss within the upscale block, the model generates three SR images $\{I_{SR_k}\}_{k=1,2,3}$ and three corresponding detail images $\{D_{SR_k}\}_{k=1,2,3}$. All these images contribute to the final loss function $L$,

$$L^{(k)} = \|I_{HR} - I_{SR_k}\| + \lambda \cdot \|D_{GT} - D_{SR_k}\|, \qquad (2)$$

$$L = \sum_{k=1}^{3} w_k \cdot L^{(k)}. \qquad (3)$$

In our experiments, after completing RUDP, we assign weights of 1, 3, and 10 to the losses associated with the SR and detail images at each successive stage, that is, $\{w_1, w_2, w_3\} = \{1, 3, 10\}$.

## 3.4 | Our Model Description

We categorize various models into CNN-based and attention-based types, and within each category, we compare the effects of our proposed methods. In this subsection, we outline the architectures of our models used for the experiments presented in Section 4.

### 3.4.1 | CNN-Based Model

For the CNN-based model, we design a new architecture, LaUD, that incorporates the LP-based detail loss and RUDP. The model has sufficient depth but remains simple, without incorporating techniques beyond our two methods. Figure 2 shows the overall structure of LaUD. Our model consists of three main blocks: a feature extraction block, an upscale block, and a downscale block. In all blocks, we utilize LeakyReLU as the activation function. Unless stated otherwise, all convolution layers have a kernel size of $3 \times 3$.

**Feature extraction block.** We construct the feature extraction block using only residual blocks (RB) and skip connections. For an LR input image $I_{LR} \in \mathbb{R}^{h \times w \times 3}$ of size $h \times w$, the shallow feature map $H_0 \in \mathbb{R}^{h \times w \times 256}$ is extracted by a convolution layer. Several residual blocks with skip connection, then extract deeper features,

$$H_n = H_{n-1} + RB_n(H_{n-1}), n = 1, 2, \ldots, N, \qquad (4)$$

where $RB_n$ is the $n$th residual block. We set $N = 4$ to ensure a sufficient model depth and fix the dimensions of each $H_n$ to $h \times w \times 256$.

**Upscale block.** The final feature map $H_N$ is provided to the upscale block. The upscale block generates both the upscaled feature map $H_{U_k} \in \mathbb{R}^{(h \times s) \times (w \times s) \times 256}$ and the detail feature map $H_{D_k} \in \mathbb{R}^{(h \times s) \times (w \times s) \times 256}$, where $k$ indicates the sequential order of upscaling within RUDP and $s$ denotes the upscaling factor. These two feature maps are then combined through element-wise addition to form the SR feature map $H_{SR_k} \in \mathbb{R}^{(h \times s) \times (w \times s) \times 256}$, similar to the reconstruction process in LP. Hence, for $k = 1, 2, \ldots, K$,

$$H_{U_k} = f(f^{\uparrow}(H_N)), \qquad (5)$$

$$H_{D_k} = f(f(f(H_{U_k}))), \qquad (6)$$

$$H_{SR_k} = H_{U_k} + H_{D_k}, \qquad (7)$$

where $f$ denotes a convolution layer and $f^{\uparrow}$ denotes a deconvolution layer (via a transposed convolution) used for upscaling. As a result, our upscale block produces two feature maps: the detail feature map $H_{D_k}$ and the SR feature map $H_{SR_k}$. In the upscale block illustrated in Figure 2, the upper output arrow represents $H_{D_k}$, and the lower output arrow represents $H_{SR_k}$. Each of these feature maps forms a distinct loss.

Unlike the back-projection upsampling methods presented in [21, 46], our upscale block generates the SR feature map using a deconvolution layer followed by a few convolution layers, thus avoiding complex structures involving multiple processing steps. Although this simplified design might raise concerns regarding performance, it has been observed that our detail loss effectively addresses potential shortcomings, enabling the model to generate appropriate SR feature maps even with this simple architecture.

**Downscale block and repetition.** To downsample the SR feature map within RUDP, we employ a convolution layer with a stride for downsampling, followed by another convolution layer:

$$H_{Down_k} = f(f_{\downarrow}(H_{SR_k})), \ k = 1, .., K - 1, \qquad (8)$$

where $f_{\downarrow}$ indicates a convolution layer with downsampling. The generated $H_{Down_k} \in \mathbb{R}^{h \times w \times 256}$ is concatenated with the LR input image $I_{LR}$ and the LR feature maps $\{H_{Down_l}\}_{l=1,\ldots,k-1}$ obtained from previous RUDP steps, and then used as input to the subsequent feature extraction block. Through this mechanism, the feature extraction block can extract more diverse information for the next SR image by referencing SR features generated in earlier steps. In LaUD, we set $K = 3^{\dagger}$. As a result, LaUD generates the detail feature maps $\{H_{D_k}\}_{k=1,2,3}$ and the SR feature maps $\{H_{SR_k}\}_{k=1,2,3}$.

---

$\dagger$ This value was selected based on empirical observations; see the Appendix for details.

**Resulting images and loss function.** The above SR and detail feature maps must be converted into RGB format at the end. We achieve this conversion using respective final $1 \times 1$ convolution layers, hereafter referred to as ToRGB layers.

The complete loss function used to train LaUD is defined in Equations (2) and (3). In these equations, $I_{SR_k} \in \mathbb{R}^{(h \times s) \times (w \times s) \times 3}$ and $D_{SR_k} \in \mathbb{R}^{(h \times s) \times (w \times s) \times 3}$ denote the outputs of LaUD, obtained by converting the SR feature map $H_{SR_k}$ and detail feature map $H_{D_k}$ through their respective ToRGB layers. As mentioned in Section 3.2, the ground-truth detail image $D_{GT}$ corresponds exactly to the largest detail image obtained during the LP process (specifically, $\Delta I_1$ in Figure 1). We adopt the $l_1$-norm for the loss function, as it effectively mitigates the smoothing effects and shows a notable capability in image restoration [47].

### 3.4.2 | Attention-Based Models

For the attention-based model, we aim to demonstrate that our methodology integrates effectively without disrupting the existing attention mechanisms. Therefore, we apply our methods to some existing attention-based models and compare the results with those of the original models. These experiments show that our LP-based detail control loss is not confined to CNN but is also effective with attention mechanisms.

As mentioned in Section 3.2, the LP-based detail loss can be implemented with minor modifications to the output part of a model. However, some models require significant structural changes to incorporate our RUDP, which helps to enhance the effectiveness of detail loss. Since these changes may not provide a valid basis for a fair comparison, only the LP-based detail loss is applied to such models.

**Choice of base models.** We initially aimed to include SOTA models for thorough evaluation; however, our computing resources limited the experiments to models requiring less memory during training. Although we did not test our methods on all models, we demonstrated their effectiveness in attention-based models based on the trends observed in the selected ones.

In choosing the base models, the following criteria are considered: (i) Availability of the authors' code to allow reproduction. (ii) Feasibility of training within our resource constraints. (iii) Demonstration of sufficiently high performance. (iv) Distinct attention approach in each model. As a result, three models are selected–ABPN [21], HAN [22], and DRLN [19].

To isolate the effects of our methods, we reproduce the original model and compare it with the version to which our methods are applied. The reproduction of each model is carried out using the code provided in its GitHub repository. We primarily use the hyperparameters specified in the original papers, and for any details not mentioned, we follow the default settings in the code. When applying our methods, all hyperparameters are kept identical to those in the reproduced version.

**Application of our methods.** To apply our detail loss, the model must generate a detail image separately from the SR image. Hence, the output part of each model requires some modifications. Here, we briefly describe our modification for each model. More details are provided in the Appendix.

ABPN has an iterative "up- and down-sample" structure similar to RUDP but generates an SR image by aggregating all upscaled features produced during the mid-process. Consequently, we cannot directly integrate RUDP; instead, we replace the "up- and down-sample" blocks of the original ABPN with our upscale and downscale blocks from LaUD, minimizing changes to the existing model methodology. Since the attention mechanism in ABPN operates on features after downsampling, our modification enables the model to handle detail features without altering the attention mechanism.

HAN employs a structure in which layer and channel-spatial attentions are applied after feature extraction by using residual channel attention blocks. Since incorporating RUDP into HAN would require significant modifications to the model's structure, we apply only the LP-based detail control. Consequently, we conduct experiments by adding a block, structured identically to the original upsampler of HAN, to generate the detail image.

DRLN applies attention within dense residual Laplacian modules, which overlap several times to form a cascading block. The entire model is composed of several such cascading blocks. Therefore, we integrate RUDP by inserting the upscale and downscale blocks from LaUD between some of these cascading blocks. This enables us to apply detail loss and RUDP while keeping the original attention structure.

## 4 | EXPERIMENTS

In this section, we begin by comparing the performance of LaUD, our CNN-based model, with SOTA model. We then present the impact of our methods when combined with attention-based models. The complete details of how we train and evaluate LaUD and attention-based models can be found in the Appendix. All implementation codes are available at https://github.com/sangjun7/LaUD.

### 4.1 | Performance of Model LaUD

Table 1 presents a comparison between LaUD and SOTA models. Following the standard conventions in the field, we conduct experiments using three datasets: Set5 [52], Set14 [53],

**TABLE 1** Quantitative comparison with state-of-the-art models on benchmark datasets. The best values among attention-based models are highlighted in red, and the best values among CNN-based models are highlighted in blue. The last bold-faced row in the Model column within each scale is our LaUD model.

| Scale | Model | Base | Set5 | | Set14 | | BSD100 | |
|---|---|---|---|---|---|---|---|---|
| | | | PSNR (dB) | SSIM | PSNR | SSIM | PSNR | SSIM |
| 2× | EDSR [14] | CNN | 38.11 | 0.9602 | 33.92 | 0.9195 | 32.32 | 0.9013 |
| | MWCNN [17] | | 37.91 | 0.9600 | 33.70 | 0.9182 | 32.23 | 0.8999 |
| | D-DBPN [46] | | 38.09 | 0.9600 | 33.85 | 0.9190 | 32.27 | 0.9000 |
| | HBPN [48] | | 38.13 | 0.961 | 33.78 | 0.921 | 32.33 | 0.902 |
| | MICU [49] | | 37.93 | 0.9601 | 33.63 | 0.9170 | 32.17 | 0.8987 |
| | RCAN [50] | Attention | 38.27 | 0.9614 | 34.12 | 0.9216 | 32.41 | 0.9027 |
| | DRLN [19] | | 38.27 | 0.9616 | 34.28 | 0.9231 | 32.44 | 0.9028 |
| | HAN [22] | | 38.27 | 0.9614 | 34.16 | 0.9217 | 32.41 | 0.9027 |
| | EDT-B [23] | | 38.63 | 0.9632 | 34.80 | 0.9273 | 32.62 | 0.9052 |
| | SwinFIR [25] | | 38.65 | 0.9633 | 34.93 | 0.9276 | 32.64 | 0.9054 |
| | HAT-L [26] | | 38.91 | 0.9646 | 35.29 | 0.9293 | 32.74 | 0.9066 |
| | CPAT [51] | | 38.68 | 0.9633 | 34.91 | 0.9277 | 32.64 | 0.9056 |
| | **LaUD** | CNN | 38.45 | 0.9625 | 34.65 | 0.9256 | 32.54 | 0.9042 |
| 4× | EDSR [14] | CNN | 32.46 | 0.8968 | 28.80 | 0.7876 | 27.71 | 0.7420 |
| | MWCNN [17] | | 32.12 | 0.8941 | 28.41 | 0.7816 | 27.62 | 0.7355 |
| | D-DBPN [46] | | 32.47 | 0.8980 | 28.82 | 0.7860 | 27.72 | 0.7400 |
| | HBPN [48] | | 32.55 | 0.900 | 28.67 | 0.785 | 27.77 | 0.743 |
| | MICU [49] | | 32.21 | 0.8945 | 28.65 | 0.7820 | 27.57 | 0.7359 |
| | RCAN [50] | Attention | 32.63 | 0.9002 | 28.87 | 0.7889 | 27.77 | 0.7436 |
| | DRLN [19] | | 32.63 | 0.9002 | 28.94 | 0.7900 | 27.83 | 0.7444 |
| | ABPN [21] | | 32.69 | 0.900 | 28.94 | 0.789 | 27.82 | 0.743 |
| | HAN [22] | | 32.64 | 0.9002 | 28.90 | 0.7890 | 27.80 | 0.7442 |
| | EDT-B [23] | | 33.06 | 0.9055 | 29.23 | 0.7971 | 27.99 | 0.7510 |
| | SwinFIR [25] | | 33.20 | 0.9068 | 29.36 | 0.7993 | 28.03 | 0.7520 |
| | HAT-L [26] | | 33.30 | 0.9083 | 29.47 | 0.8015 | 28.09 | 0.7551 |
| | CPAT [51] | | 33.19 | 0.9069 | 29.34 | 0.7991 | 28.04 | 0.7527 |
| | **LaUD** | CNN | 32.82 | 0.9021 | 29.06 | 0.7939 | 27.89 | 0.7472 |
| 8× | EDSR [14] | CNN | 26.96 | 0.7762 | 24.91 | 0.6420 | 24.81 | 0.5985 |
| | D-DBPN [46] | | 27.21 | 0.7840 | 25.13 | 0.6480 | 24.88 | 0.6010 |
| | HBPN [48] | | 27.17 | 0.785 | 24.96 | 0.642 | 24.93 | 0.602 |
| | RCAN [50] | Attention | 27.31 | 0.7878 | 25.23 | 0.6511 | 24.98 | 0.6058 |
| | DRLN [19] | | 27.36 | 0.7882 | 25.34 | 0.6531 | 25.01 | 0.6057 |
| | ABPN [21] | | 27.25 | 0.786 | 25.08 | 0.638 | 24.99 | 0.604 |
| | HAN [22] | | 27.33 | 0.7884 | 25.24 | 0.6510 | 24.98 | 0.6059 |
| | **LaUD** | CNN | 27.51 | 0.7882 | 25.34 | 0.6569 | 25.04 | 0.6102 |

and BSD100 [54]. We evaluate the Peak Signal to Noise Ratio (PSNR) and Structural Similarity Index Measure (SSIM) values for 2×, 4×, and 8× upscaling. However, since some papers do not report 8× upscaling results, we include only the reported results for 8× upscaling. The PSNR and SSIM are calculated on the Y channel from the YCbCr space. While

PSNR and SSIM are fundamental for evaluating reconstruction accuracy, they may not fully align with human perceptual quality. We therefore conduct a complementary evaluation with various perceptual metrics in Section 5.1.

When considering the overall results in the table, LaUD consistently outperforms all CNN-based models. D-DBPN and HBPN are CNN-based models that use the back-projection method, which is similar to our RUDP. While these models perform well within CNN-based architectures, LaUD achieves better results with the LP-based detail loss. In addition, LaUD demonstrates performance comparable to attention-based models, despite employing a CNN-based architecture. It outperforms RCAN, DRLN, ABPN, and HAN across all datasets. In our view, surpassing attention-based models, which excel at feature extraction, demonstrates the effectiveness of our LP-based detail control and RUDP in extracting and utilizing features.

**TABLE 2** Performance of attention-based models with our detail loss or RUDP. "+" denotes the use of our methods. Specifically, we apply both detail loss and RUDP to DRLN, only detail loss to the others.

| Scale | Model | Set5 | | Set14 | | BSD100 | |
|-------|-------|------|------|-------|------|--------|------|
| | | PSNR | SSIM | PSNR | SSIM | PSNR | SSIM |
| 2× | DRLN | 38.10 | 0.9610 | 33.75 | 0.9188 | 32.24 | 0.9006 |
| | DRLN+ | 38.35 | 0.9623 | 34.49 | 0.9241 | 32.49 | 0.9039 |
| | HAN | 38.26 | 0.9616 | 34.11 | 0.9217 | 32.39 | 0.9027 |
| | HAN+ | 38.31 | 0.9617 | 34.15 | 0.9220 | 32.41 | 0.9030 |
| 4× | ABPN | 32.28 | 0.8955 | 28.67 | 0.7828 | 27.61 | 0.7379 |
| | ABPN+ | 32.60 | 0.8991 | 28.90 | 0.7882 | 27.76 | 0.7433 |

A detailed examination of the table reveals that, for 2× upscaling, LaUD achieves improvements of 0.18 dB on Set5 and 0.37 dB on Set14 compared to DRLN, an attention-based model that employs the concept of LP, similar to our model. Compared to the previously mentioned models, D-DBPN and HBPN, LaUD demonstrates significant performance improvements, achieving gains of at least 0.32 dB on Set5 and 0.8 dB on Set14. A similar trend in all the above comparisons is also observed in terms of SSIM results. We interpret these results as suggesting that our model effectively restores high-frequency data for super-resolution by incorporating the LP-based detail loss and RUDP.

The influence of detail control is maintained even as the scaling increases. For instance, at 8× upscaling, LaUD achieves PSNR values of 27.51 dB on Set5, 25.34 dB on Set14, and 25.04 dB on BSD100, outperforming all other models on these datasets. Since an LR image in the 8× SR task contains significantly less information than an HR image, it is challenging to extract an appropriate feature map from the LR image using only the loss computed on the SR image. Given this difficulty, our results suggest that employing detail loss plays an important role in capturing information in the 8× upscaled features, which might otherwise remain missing.

## 4.2 | Application to Attention-Based Models

This subsection presents the results of applying our LP-based detail loss to attention-based models. As outlined in Section 3.4.2, we selected three models: DRLN, HAN, and ABPN. Depending on the structure of each model, we applied either the detail loss alone or together with RUDP (application indicated with + in model names). A detailed explanation of the model structure is provided in the Appendix.

Table 2 presents our experimental results. All original models were reproduced using the official code provided in their respective papers. Although the reproduced results did not exactly match the reported values, a fair comparison remains valid, as both the original models and those incorporating our methods were trained under identical conditions and within the same environment.

Among the three models, DRLN significantly improves performance by applying both our LP-based detail loss and RUDP. For Set5, the reproduced DRLN achieves 38.10 dB in PSNR, whereas the model incorporating our methods achieves 38.35 dB in PSNR, demonstrating an improvement of 0.25 dB. For Set14, the difference between DRLN and DRLN+ is even larger, at 0.74 dB. The SSIM metrics also consistently exhibit improvements for DRLN+. DRLN requires slight modifications to incorporate RUDP due to its multi-block structure. Consequently, its existing attention mechanism is preserved and effectively integrated with our two methods, resulting in positive synergy.

We only apply the LP-based detail loss to HAN, as adding RUDP poses a risk of significantly altering the structure. This results in slight improvements for PSNR, such as an increase from 38.26 dB to 38.31 dB on Set5 and from 32.39 dB to 32.41 dB on BSD100. Although the difference is modest, incorporating the detail loss consistently improves the performance of HAN across all datasets. Since HAN utilizes RCAN as a pretrained model, the sub-pixel convolution for the SR image is also pretrained. However, the component responsible for producing the detail image used in our detail loss needs to be trained from scratch with a small learning rate. This likely explains why the PSNR value does not show a more significant difference.

**TABLE 3** Quantitative results with additional image-quality metrics. We consider six metrics: full-reference (FSIM, DISTS, LPIPS), no-reference (NIQE, CLIP-IQA), and distribution-level (FID). All results are for 2× SR on benchmark datasets, using a subset of models from Table 1. ↑ means "higher is better," while ↓ means "lower is better." The best values among attention-based models are highlighted in red, and the best values among CNN-based models are highlighted in blue. The last row in bold in the Model column within each dataset corresponds to our LaUD model.

| | Model | Base | FR | | | NR | | FID (↓) |
|---|---|---|---|---|---|---|---|---|
| | | | FSIM (↑) | DISTS (↓) | LPIPS (↓) | NIQE (↓) | CLIP-IQA (↑) | |
| Set5 | D-DBPN | CNN | 0.9808 | 0.0785 | 0.0539 | 5.2655 | 0.7651 | 21.3355 |
| | HBPN | | 0.9804 | 0.0794 | 0.0550 | 5.3195 | 0.7694 | 25.0403 |
| | DRLN | Attention | 0.9812 | 0.0780 | 0.0533 | 5.2579 | 0.7752 | 23.3138 |
| | SwinFIR | | 0.9828 | 0.0736 | 0.0492 | 5.1583 | 0.7813 | 18.9931 |
| | HAT-L | | 0.9838 | 0.0690 | 0.0474 | 5.1228 | 0.7860 | 16.6948 |
| | **LaUD** | CNN | 0.9817 | 0.0758 | 0.0511 | 5.1398 | 0.7806 | 21.5852 |
| Set14 | D-DBPN | CNN | 0.9835 | 0.0795 | 0.0890 | 5.1008 | 0.6753 | 11.6767 |
| | HBPN | | 0.9819 | 0.0826 | 0.0905 | 5.1970 | 0.6758 | 13.4095 |
| | DRLN | Attention | 0.9837 | 0.0814 | 0.0894 | 5.1200 | 0.6783 | 12.5353 |
| | SwinFIR | | 0.9852 | 0.0770 | 0.0807 | 5.0196 | 0.6877 | 10.6579 |
| | HAT-L | | 0.9856 | 0.0752 | 0.0774 | 4.9694 | 0.6879 | 9.5935 |
| | **LaUD** | CNN | 0.9844 | 0.0780 | 0.0834 | 5.0225 | 0.6821 | 10.5419 |
| BSD100 | D-DBPN | CNN | 0.9483 | 0.1014 | 0.1432 | 4.9832 | 0.6873 | 13.8032 |
| | HBPN | | 0.9475 | 0.1027 | 0.1425 | 5.0583 | 0.6963 | 14.3469 |
| | DRLN | Attention | 0.9488 | 0.1037 | 0.1389 | 5.0186 | 0.6843 | 13.7294 |
| | SwinFIR | | 0.9511 | 0.1008 | 0.1316 | 4.9405 | 0.6989 | 12.7832 |
| | HAT-L | | 0.9519 | 0.0995 | 0.1301 | 4.8946 | 0.6929 | 12.3747 |
| | **LaUD** | CNN | 0.9498 | 0.1005 | 0.1352 | 4.9083 | 0.6951 | 13.0558 |

For ABPN+, we conducted experiments on 4× upscaling, following the original ABPN setup. With the addition of our detail loss, the PSNR improves by 0.32 dB on Set5, 0.23 dB on Set14, and 0.15 dB on BSD100. Moreover, the improvements in SSIM are the most significant among these experiments. Generating fine details is more challenging in the process of 4× upscaling compared to 2× upscaling. These observations imply that incorporating the LP-based detail loss allows ABPN+ to more effectively extract high-frequency detail that ABPN does not sufficiently capture, leading to a significant improvement in model performance.

In summary, across all models and datasets, combining the attention-based model with our detail loss leads to performance improvements in PSNR and SSIM metrics. This shows that our LP-based detail loss is not limited to CNN structures, but can be effectively integrated with an attention mechanism to enhance a model.

## 5 | FURTHER ANALYSIS OF LAUD

We now present our analysis for LaUD to validate the effectiveness of the LP-based detail control and RUDP. We expect that a similar analysis could be conducted when applying our methods to existing attention-based models; however, we focus our analysis on LaUD to specifically examine the effects of our methods independently of any influence from attention mechanisms. In addition to the analysis presented in this section, we provide additional ablation studies and feature map examples in the Appendix and on our GitHub, along with examples of output images and additional visual comparisons involving LaUD.

**TABLE 4**  Ablation for the LP-based detail control and RUDP (for 2×). When RUDP is applied, the total loss is the weighted sum of the losses for each image generated during RUDP (see Section 3.3).

| Model Name | RUDP | Detail Control | Set5 | | Set14 | | BSD100 | |
|---|---|---|---|---|---|---|---|---|
| | | | PSNR | SSIM | PSNR | SSIM | PSNR | SSIM |
| M1 | X | X | 38.0857 | 0.9613 | 33.9680 | 0.9205 | 32.2910 | 0.9015 |
| M2 | X | O | 38.2841 | 0.9619 | 34.2761 | 0.9224 | 32.4013 | 0.9029 |
| M3 | O | X | 38.3154 | 0.9620 | 34.6050 | 0.9250 | 32.4888 | 0.9037 |
| M4 | O | O | 38.4237 | 0.9625 | 34.7677 | 0.9256 | 32.5504 | 0.9045 |

## 5.1 | Results on Additional Image-Quality Metrics

As discussed earlier, LaUD is evaluated against SOTA models using PSNR and SSIM. While these are standard reconstruction metrics, they may not fully capture perceptual quality, so we incorporate complementary metrics.

Image-quality assessment metrics are commonly grouped into: full-reference (FR), which compare reconstructions with ground-truth images; no-reference (NR), which assess quality without a reference; and distribution-level metrics, which compare sets of generated and ground-truth images. We consider six representative metrics. For FR, we employ FSIM [55], LPIPS [56], which emphasizes perceptual similarity via deep features, and DISTS [57], a distance that balances structural and textural similarity. For NR, we adopt NIQE [58] and CLIP-IQA [59]. For distribution-level, we calculate FID [60], which measures distributional realism rather than per-image similarity. In these experiments, we evaluate a subset of models from Table 1 for which official code and pretrained weights are publicly available. For each model, we regenerate SR images using the released code and compute all metrics via `pyiqa`. All experiments focus on the 2× SR process.

Table 3 summarizes the results. In the metric header, ↑ means "higher is better," while ↓ means "lower is better." Our LaUD outperforms D-DBPN, HBPN, and DRLN across all FR and NR metrics, with the sole exception of CLIP-IQA for HBPN on BSD100. In terms of FID, LaUD also exceeds these three models, except for D-DBPN on Set5. Relative to Swin-FIR, LaUD is generally comparable overall, though SwinFIR is slightly ahead in more cases. LaUD surpasses SwinFIR in DISTS on BSD100, in NIQE on Set5 and BSD100, and in FID on Set14, achieving the second-best overall performance.

When focusing on specific metrics, LaUD ranks just below HAT-L and SwinFIR in FSIM. LaUD again ranks behind HAT-L and SwinFIR in terms of LPIPS, scoring 0.0834 on Set14 and 0.1352 on BSD100. For DISTS, LaUD is comparable to SwinFIR with scores of 0.0758 on Set5 and 0.0780 on Set14, and surpasses it with 0.1005 on BSD100. Consistent with the FR results, the NR metrics NIQE and CLIP-IQA indicate that LaUD is competitive with SwinFIR. Particularly for NIQE, LaUD records 5.0225 on Set14, which is close to SwinFIR's 5.0196. On BSD100, LaUD achieves 4.9083, surpassing SwinFIR. Considering FID, LaUD achieves 10.5419 on Set14 and 13.0558 on BSD100. Overall, these results suggest that LaUD produces perceptually faithful SR images, as our LP-based detail loss effectively encourages the model to capture and enhance high-frequency details.

## 5.2 | Ablation Study on LaUD

We conduct ablation studies to assess the impact of our LP-based detail control and RUDP. For comparison, we examine four scenarios (named M1 through M4) depending on whether they employ RUDP and the LP-based detail control. We train each model for the four scenarios only once on ImageNet 2012 [61]. To minimize the influence of model complexity, we adjust the architectures to have a similar number of parameters across models. We exclusively focus on 2× upscaling, applying consistent training hyperparameters across all experiments.

As shown in Table 4, the results demonstrate the impact of the LP-based detail loss and the synergistic effect when RUDP is applied simultaneously. In particular, when comparing M1 and M2, as well as M3 and M4, the models trained with the detail loss consistently outperform their counterparts in terms of both PSNR and SSIM. This indicates that guiding the model with detail loss is an effective approach, especially in enhancing the qualities of the SR image. The addition of RUDP also enhances the performance of the models. For instance, the PSNR of model M1 on Set5 is 38.0857 dB, whereas that of model M3, which uses only RUDP, is 38.3154 dB. RUDP facilitates model training by utilizing images from intermediate layers as losses, guiding the model to generate accurate SR images at those stages. This approach enables the model to extract more relevant features through RUDP and progressively refine the SR image in subsequent steps.

Notably, the model that applies all methods achieves the highest performance, with PSNR values of 38.4237 dB on Set5, 34.7677 dB on Set14, and 32.5504 dB on BSD100. This model, M4, shows a significant improvement over M1, achieving gains of approximately 0.34 dB on Set5, 0.80 dB on Set14, and 0.26 dB on BSD100. These substantial differences indicate that the LP-based detail loss and RUDP yield a synergistic effect. While the detail loss appears to guide the model to focus more on high-frequency components, potential deficiencies can be mitigated by re-extracting features from the SR image produced in the previous step through RUDP.

## 5.3 | SR Feature Map Analysis: Impact of Our Methods

We analyze the output feature maps of our CNN-based model to illustrate the effectiveness of the proposed LP-based detail loss and RUDP. Figure 3 illustrates the SR feature maps generated by LaUD, comparing cases with and without our detail loss. Note that without our detail loss, the loss function in Equation (1) becomes $\|I_{HR} - I_{SR}\|$. Before the final ToRGB layer that converts the feature maps into RGB channels, the feature maps consist of 256 channels. However, in this figure, we focus on only the last 16 channels to highlight the changes more clearly. The complete set of channels for this image and the feature map for another image can be found in the Appendix.

The figure displays two feature maps: the initial SR feature map $H_{SR_1}$ (left column) and the final SR feature map $H_{SR_3}$ (right column) in RUDP. In both cases (a) and (b), the final SR feature map has more channels exhibiting strong activation than the initial SR feature map. Additionally, the contrast between channels within each final feature map is more clearly distinguishable than in the corresponding initial feature map. These differences indicate that the upscaling and downscaling processes enable the model to enrich the features extracted for SR.

When comparing the feature maps in (a) and (b), we observe clearly that the detail loss affects the diversity of the feature maps. As shown in the feature map in (b), if there is no guidance for the model to capture high-frequency information, RUDP amplifies only a few prominent channels, while most channels remain at values with low contrast, close to neutral gray. In contrast, for (a), even in the initial SR feature map, shapes containing texture are visible across many channels. Notably, in the final feature map, this texture is further enhanced. As a result, each channel conveys distinct and clear texture information. This difference highlights the importance of guiding high-frequency information by the detail loss in SR tasks. It also demonstrates that RUDP, when combined with the LP-based detail control, provides significant benefits by effectively amplifying diverse information for high-frequency components.

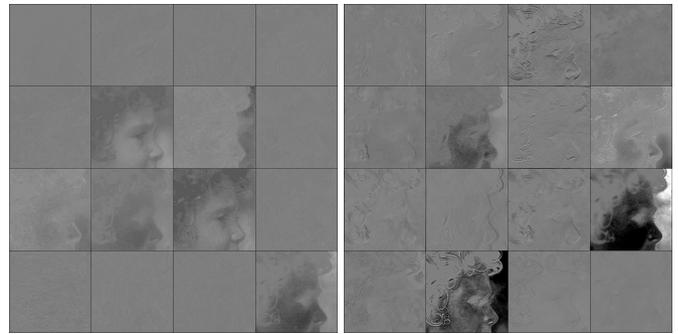

(a) From LaUD.

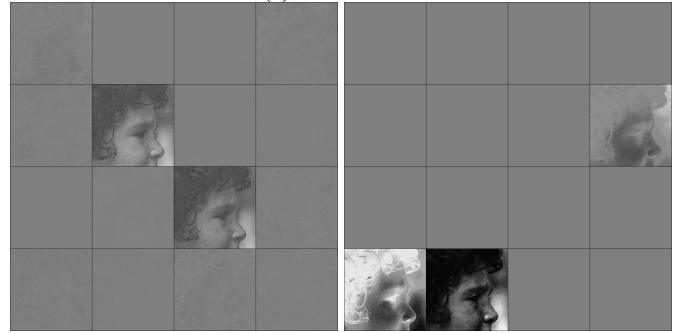

(b) From LaUD without detail loss.

**FIGURE 3** Parts of the SR feature map (for 2×): (a) from LaUD, and (b) from LaUD without detail loss. In each row, the left image shows the result after the initial upscaling step, while the right image shows the result after the final upscaling in RUDP.

## 5.4 | Feature Map Comparison in LaUD

As explained in Section 3.4.1, when designing LaUD, we derive a detail feature map $H_{D_k}$ from the upscaled feature map $H_{U_k}$, and combine them to form the SR feature map $H_{SR_k}$ (see Equation (7)). Without applying our detail loss, this structure, which generates $H_{SR_k}$, would function merely as a skip connection, adding slightly adjusted $H_{D_k}$ to $H_{U_k}$. However, our detail loss imposes constraints on $H_{D_k}$, distinguishing it from $H_{SR_k}$. As a result, $H_{D_k}$ exhibits distinct characteristics compared to $H_{SR_k}$ and $H_{U_k}$. Then $H_{SR_k}$ is generated by updating $H_{U_k}$ to incorporate new information present in $H_{D_k}$. When we examine the feature maps, $H_{D_k}$ generally captures information about fine details such as boundaries and textures. Hence, when combined with $H_{U_k}$, $H_{D_k}$ enriches the information in $H_{U_k}$ and helps to adjust values that are excessively flat or overly emphasized.

We illustrate the above analysis by presenting the feature map results of a sample image from the Set5 dataset. Figure 4 displays selected regions of the feature maps generated during the 2× SR process of LaUD. All feature maps shown are generated during the final upscaling step of RUDP. Examining this figure, we observe that the upscaled feature map

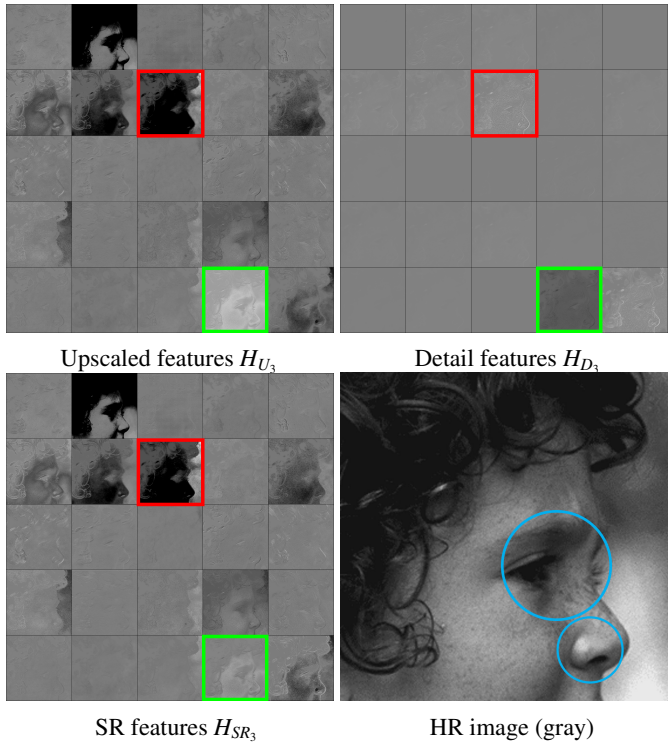

Upscaled features $H_{U_3}$      Detail features $H_{D_3}$

SR features $H_{SR_3}$      HR image (gray)

**FIGURE 4** Selected regions of three different feature maps generated during the $2\times$ SR process of LaUD for a sample image from Set5. The bottom-right image is the ground-truth HR image shown in grayscale for reference.

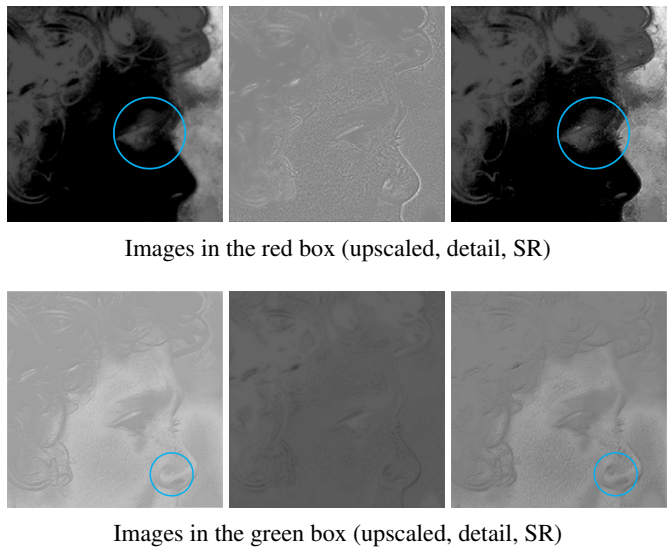

Images in the red box (upscaled, detail, SR)

Images in the green box (upscaled, detail, SR)

**FIGURE 5** Enlarged views of the regions marked by red (top row) and green (bottom row) boxes in Figure 4. Each row shows the upscaled, detail, and SR features.

predominantly retains low-frequency information, such as general object structures, and contains features exhibiting varying levels of contrast. On the other hand, the detail feature map frequently highlights distinct boundaries or textures not captured in the upscaled feature map, despite showing less variation overall.

Specifically, we now focus on a detailed comparison of the two cases highlighted by the red and green boxes. Figure 5 presents an enlarged view of the features within the red and green boxes from Figure 4. Each row, from left to right, corresponds to the upscaled, detail, and SR features, respectively.

In the case of the images in the red box, the detail feature reveals prominent boundaries and textures. As a result, when this detail feature is combined with the upscaled feature to form the SR feature, the insufficient high-frequency information of the upscaled feature is effectively reinforced. For example, examining the images in the first row of Figure 5 and the HR image (in grayscale) from Figure 4, we observe that, compared to the upscaled feature, the SR feature exhibits more distinct facial profile and enhanced emphasis around the eyes, with clearly defined eyelashes (the circled area).

Next, for the images in the green box, the detail feature serves a different role compared to the red box. We observe that the upscaled feature in the green box generally exhibits rather flat values. However, this flatness is somewhat alleviated by incorporating the detail feature. Consequently, the SR feature demonstrates greater value curvature and appears to capture a more dynamic appearance, particularly enhancing fine textures around the nose crease (circled area in the second row of Figure 5).

## 5.5 | Visual Comparisons

We also provide the visual comparison in Figure 6. We compare the SR images produced by LaUD with those generated by other SOTA models on Urban100 [62]. Urban100 consists of images where structural information is crucial, such as buildings with numerous windows or spiral staircases converging to a point. This comparison allows us to evaluate whether the model can accurately identify and reproduce repetitive structures down to the fine details in the SR process.

The figure presents the results of D-DBPN, DRLN, Swin-FIR, HAT-L, and our LaUD across two images. Additional examples are reported in the Appendix. Although SwinFIR and HAT-L generally achieve higher PSNR values than LaUD (as shown in Table 1), our LaUD demonstrates superior performance to SwinFIR and HAT-L in recovering certain structural details, as shown in Figure 6. We illustrate these structures through visual comparisons below, highlighting the effects of incorporating our LP-based detail loss and RUDP.

In the first image, a closer inspection of the patches reveals differences in the wall's detailed texture. Our LaUD achieves the highest PSNR value, with 34.5614 dB. In our view, LaUD produces an image closer to the HR image by generating a texture that resembles dust along the diagonal structural line located in the lower-left part of the image, which separates two adjacent window panels. In contrast, D-DBPN, DRLN, Swin-FIR, and HAT-L produce images that emphasize cleaner lines in this area, and thus fail to adequately represent finer details, such as dust or line width. From this result, we think that our detail loss allows the model to focus more effectively on these intricate textures.

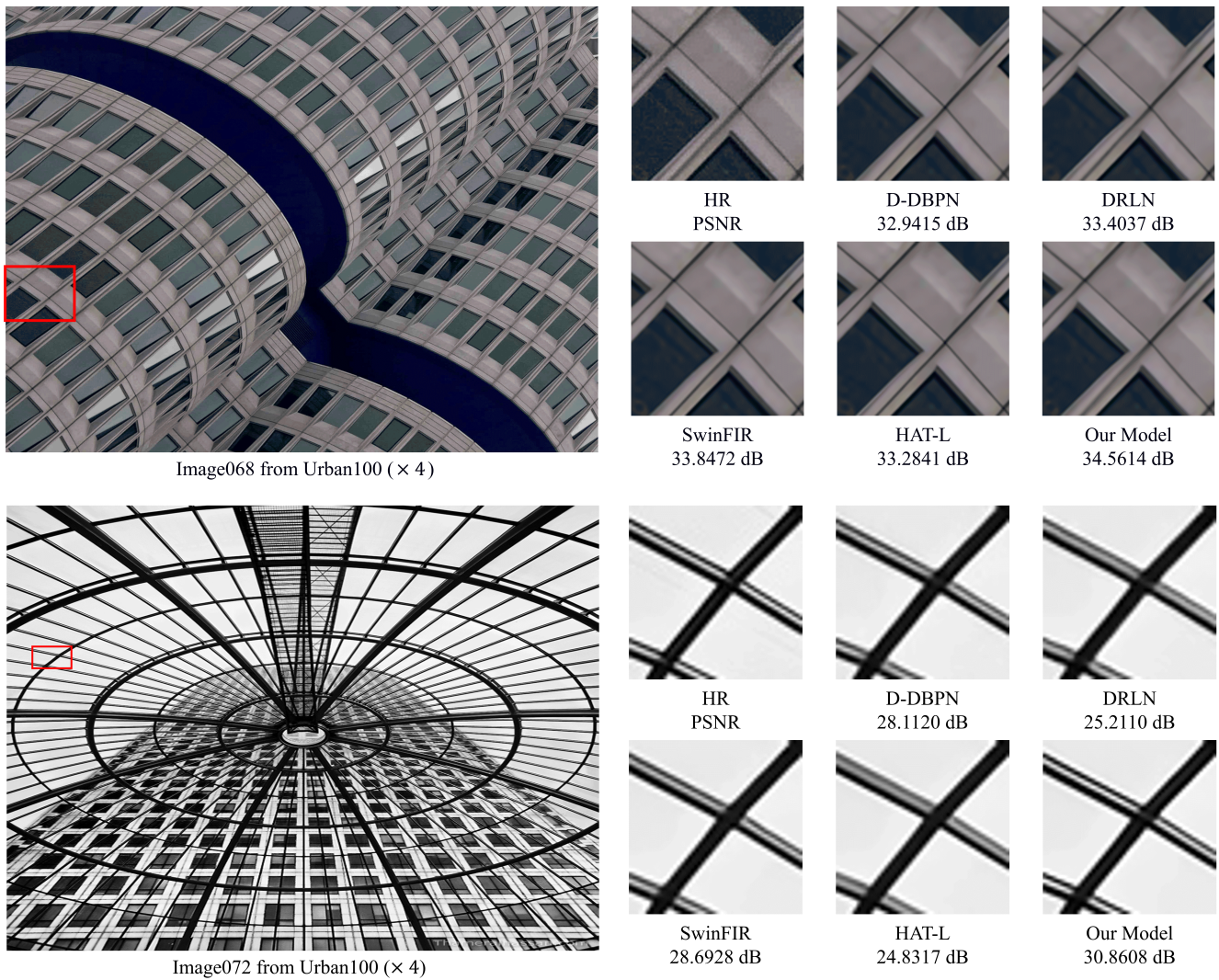

**FIGURE 6** Visual comparison for 4× SR on Urban100. The patches for comparison are marked with red boxes in the original images. The PSNR values are calculated based on the patches.

The second image, "Image072," features a pattern of circular lines. The enlarged image in the red box highlights the area where the lines intersect. All models render these lines without distortion. However, when examining the pair of parallel diagonal lines running from the upper-left to the lower-right region of the patch, our model delineates the boundary between the black bars and adjacent bright regions more clearly than other models. This enhanced separation gives the lines a sharper appearance. We attribute this result to the model's ability to capture boundaries through the LP-based detail loss and to enhance pixels via RUDP. As a result, our LaUD achieves the best PSNR value in this patch, surpassing all other models by more than 2 dB.

To demonstrate improvements in detail compared to models within the same CNN category, we provide the additional visual comparisons in Figure 7. We compare the SR images generated by LaUD with those produced by other CNN-based SOTA models listed in Table 1. These qualitative comparisons reinforce our quantitative results by illustrating that LaUD effectively recovers high-frequency structures and fine details.

We examine the ×4 upscaling results on Urban100 images "Image085" and "Image009," using two other CNN models,

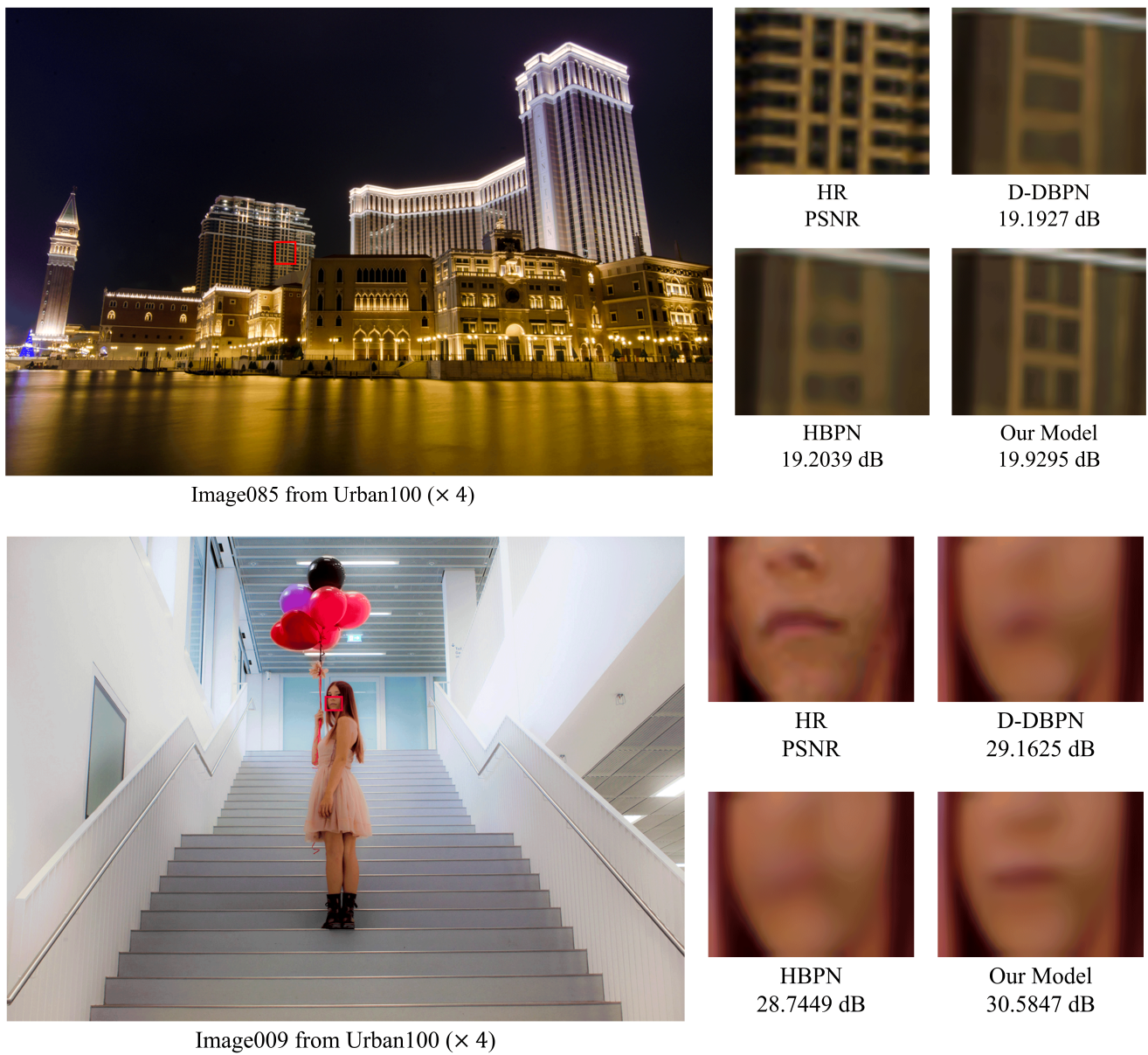

HR
PSNR

D-DBPN
19.1927 dB

HBPN
19.2039 dB

Our Model
19.9295 dB

Image085 from Urban100 (× 4)

HR
PSNR

D-DBPN
29.1625 dB

HBPN
28.7449 dB

Our Model
30.5847 dB

Image009 from Urban100 (× 4)

**FIGURE 7**  Visual comparison among CNN-based models for 4× SR on Urban100. The patches for comparison are marked with red boxes in the original images. The PSNR values are calculated based on the patches.

D-DBPN and HBPN, both of which demonstrate excellent performance, as shown in Table 1. EDSR is excluded from this comparison because its PSNR values at ×4 upscaling are not significantly different from those of D-DBPN.

The first image, "Image085" depicts a nighttime cityscape, characterized by a significant amount of high-frequency information due to sharp contrasts between brightly illuminated and dark areas. Furthermore, because luminous sources such as light bulbs often exhibit a spreading or glowing effect, there is a possibility that the model interprets this inherent blur as an intrinsic characteristic of the image during super-resolution processing, thereby increasing the difficulty of accurately reconstructing such scenes. Upon zooming into the area marked with a red box in the HR image, we observe a building window featuring distinct horizontal and vertical lines intersecting. However, D-DBPN fails to adequately reconstruct the vertical line, and HBPN generates it only faintly, making it hardly noticeable. In contrast, our model successfully reconstructs both the horizontal and vertical lines with high clarity. This difference demonstrates that LaUD effectively identifies and enhances detailed structures by accurately resolving high-frequency data even within regions of low illumination, and generates significantly less blur compared to other top-tier CNN models.

Next, "Image009" depicts a woman standing on a staircase, and we specifically focus on the region around her mouth. This area represents a very small portion of the overall image, serving as a valuable test case for evaluating a model's ability to capture fine-scale details. In examining the results, HBPN shows significant blurring around the nose and mouth, making it difficult to identify their shapes clearly. D-DBPN partially reconstructs the mouth but emphasizes only the central lip contrast, resulting in a notable deviation from the shape observed in the HR image. In contrast, our model captures the overall shape of the lips more accurately, albeit slightly straighter than in the HR image, with a length that closely matches that of the original. Additionally, our model produces subtle contrast around the nostrils, further clarifying the shape of the nose compared to the other two models. From these observations, it is evident that LaUD effectively recognizes and enhances high-frequency details within small-scale objects, thereby producing SR results that more closely resemble the HR image.

# 6 | CONCLUSIONS

In this paper, we propose a novel detail loss based on the Laplacian pyramid (LP) and a repeated upscaling and downscaling process (RUDP) for the super-resolution task. The LP-based detail loss can be used with both CNN models and attention-based architectures, as it is independent of the model's structure. In addition, when combined with RUDP, the LP-based detail loss produces a synergistic effect, significantly improving the performance. In our experiments, we construct a CNN-based model that incorporates the LP-based detail loss and RUDP. Comparison of PSNR, SSIM, and perceptual quality metrics with other state-of-the-art models, as well as ablation studies and feature map analyses, shows the effectiveness of our detail loss and RUDP in enabling the model to capture high-frequency information. Moreover, we integrate our methods into several existing attention-based models, resulting in improved performance across all of them. This demonstrates that the LP-based detail loss is effective with attention mechanisms and applicable regardless of the model structure.

## ACKNOWLEDGEMENTS

This work was supported in part by the National Research Foundation of Korea (NRF) [Grant Number RS-2021-NR058317].

## SUPPORTING INFORMATION

Additional supporting information may be found in the online version of the article at the publishers website.

## AUTHOR CONTRIBUTIONS

Sangjun Han: Conceptualization; Methodology; Software; Investigation; Formal analysis; Writingoriginal draft; Writingreview & editing.

Youngmi Hur: Conceptualization; Methodology; Supervision; Project administration; Funding acquisition; Formal analysis; Writingreview & editing.

# APPENDIX

## A APPLYING OUR METHODS TO ATTENTION-BASED MODELS

As mentioned in the main text, small modifications are required to apply our methods to the three selected attention-based models: DRLN [19], HAN [22], and ABPN [21]. We illustrate the modified version of DRLN incorporating our methods in Figure A1. For HAN and ABPN, we present both the original structures and their modified versions in Figures A2 and A3, respectively. The original structures of HAN and ABPN depicted in these figures are sourced from [21, 22].

DRLN consists of cascading blocks, each containing multiple dense residual Laplacian modules. According to the code by the authors, the original DRLN structure passes through a total of three cascading blocks, with a short skip connection after each block. After each cascading block and short skip connection, we apply the LP-based detail control and RUDP by incorporating our upscale and downscale blocks. As a result, we modify the model and obtain the structure shown in Figure A1, without altering the attention mechanism within the dense residual Laplacian modules. With the introduction of three upscale blocks, the modified DRLN generates three SR images and three corresponding detail images. The total loss is calculated as a weighted sum using these images, as in LaUD.

HAN extracts features through residual groups and then applies layer attention and channel-spatial attention to these features. Since channel attention is also present within the residual groups, introducing our RUDP during the feature extraction phase would disrupt the attention mechanism. Consequently, we decide against applying RUDP and instead integrate only our LP-based detail loss. Upon reviewing the code, we confirm that HAN utilizes RCAN as its pretrained base. This leads to insufficient training when the upscale block of LaUD replaces the original sub-pixel convolution in the upscaling stage. Therefore, we retain the sub-pixel convolution for upsampling and add an additional sub-pixel convolution to generate the detail image. The resulting modified version is illustrated in Figure A2 (b).

ABPN iteratively performs an upsampling and downsampling scheme using back-projection blocks (originally denoted as "up-/down-sampling back projection" in the ABPN paper) with an attention module applied after the downsampling stage. We replace these upsampling and downsampling back-projection blocks with the upscale and downscale blocks of LaUD, keeping the attention module intact. Since our upscale block generates both detail and SR features, defining the detail loss becomes straightforward. However, the original ABPN follows a rather complex process to produce an SR image, involving the convolution of concatenated SR and LR features, followed by the addition to the bicubic-upsampled LR image.

Replicating that pipeline for the detail part can lead to undesirable complications, such as requiring a downsampled version of the detail image. Thus, we instead use a simpler structure that aggregates detail features and generates a detail image through convolution, as shown in Figure A3 (b). Because multiple images are not produced despite applying RUDP, building the usual weighted sum of losses using multiple SR and detail images is not feasible in this case.

## B EXPERIMENTAL SETUP

### B.1 Setup for LaUD

In this section, we describe the training setting for LaUD. As mentioned in the main text, when we set three upscaling steps, the weights in each loss for SR and detail images, denoted as $\{w_k\}_{k=1,2,3}$, are set to 1, 3, and 10, respectively. Our choice of these weights was not exhaustively optimized; we simply adopted the set that yielded the best performance in our comparative experiments.

We execute two training sessions, as shown in HAT [26], SwinFIR [25], and EDT [23]: pretraining on ImageNet 2012 [61] and fine-tuning on DIV2K [63] and Flickr2K [64]. For both training sessions, the number of training epochs, initial learning rate, and learning rate schedules are based on previous studies. The model shows significant performance with pretraining alone, but fine-tuning with high-resolution images yields slight improvements.

The hyperparameters used in the two training sessions are similar to those in previous studies, such as [19, 23, 25, 26, 48]. During the pretraining stage, we resize the images in ImageNet to $224 \times 224$ and then randomly crop them to $128 \times 128$. The augmented images serve as the HR targets we aim to fit, and the LR images are generated at a size of $(128/s) \times (128/s)$ through the bicubic interpolation based on the upscaling factor $s$. We set the initial learning rate to $2 \cdot 10^{-4}$ and train models for 25 epochs. The learning rate is halved at 50%, 80%, 90%, and 96% of the total epochs.

For fine-tuning, we combine the DIV2K and Flickr2K datasets as training data. Since these images are very large, resizing significantly degrades their quality. Therefore, unlike in the pretraining stage, we perform only random cropping. However, the crop size of $128 \times 128$ used in pretraining often results in images containing no meaningful objects, negatively impacting model performance after fine-tuning. To address this, we increase the crop size to the largest dimension our computing resources could accommodate, settling on $256 \times 256$. Fine-tuning runs for 1000 epochs, starting with an initial learning rate of $10^{-5}$, which is halved at 50%, 80%, 90%, and 95% of the training progress.

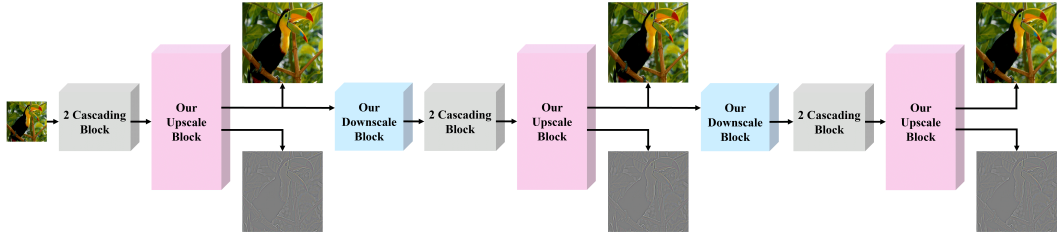

**FIGURE A1** The modified DRLN to apply our methods. The cascading block originates from the original DRLN [19], which includes its attention mechanism.

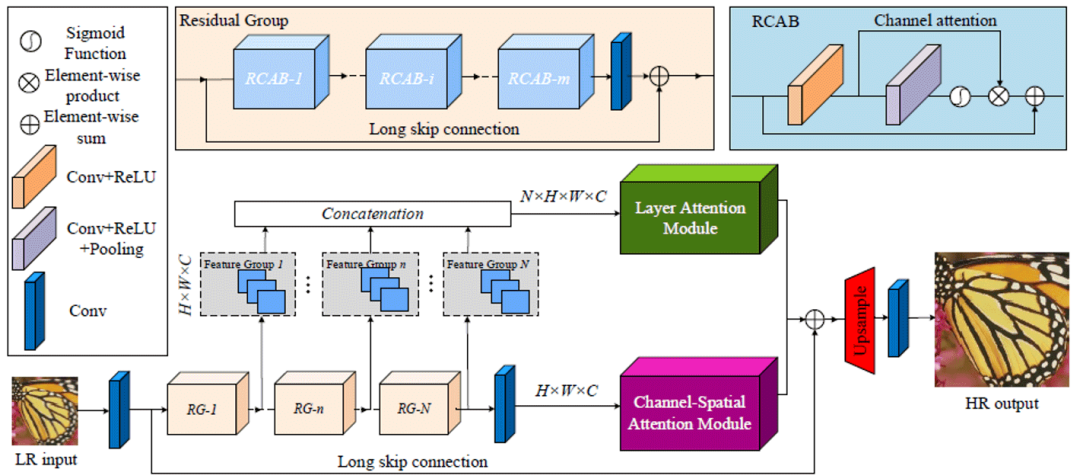

(a) The original HAN, taken from [22].

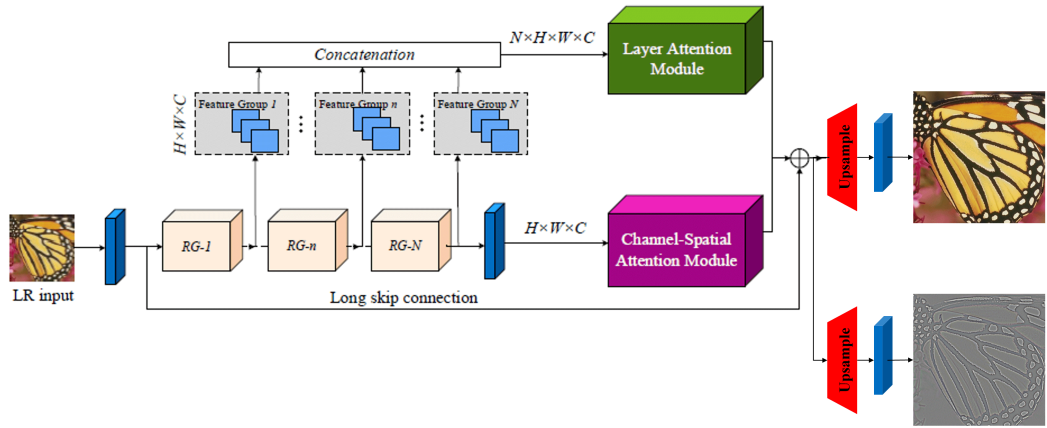

(b) The modified HAN.

**FIGURE A2** The original HAN structure and modified version to apply our methods. ((a) reproduced from [22] © 2020 Springer Nature; (b) adapted from the same source; used with permission.)

## B.2 Setup for Modified Attention-Based Model

When training the modified attention-based models incorporating our methods, most settings follow the original papers or default configurations provided by the original code. Since these settings match those used for reproducing the original models, we briefly summarize them here and recommend referring to the original papers for further details.

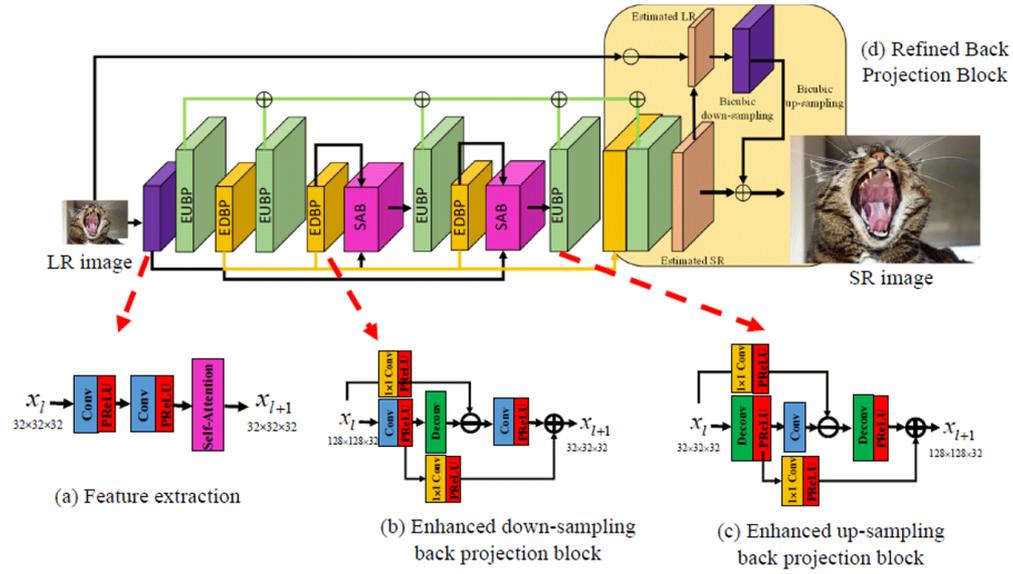

(a) The original ABPN, taken from [21].

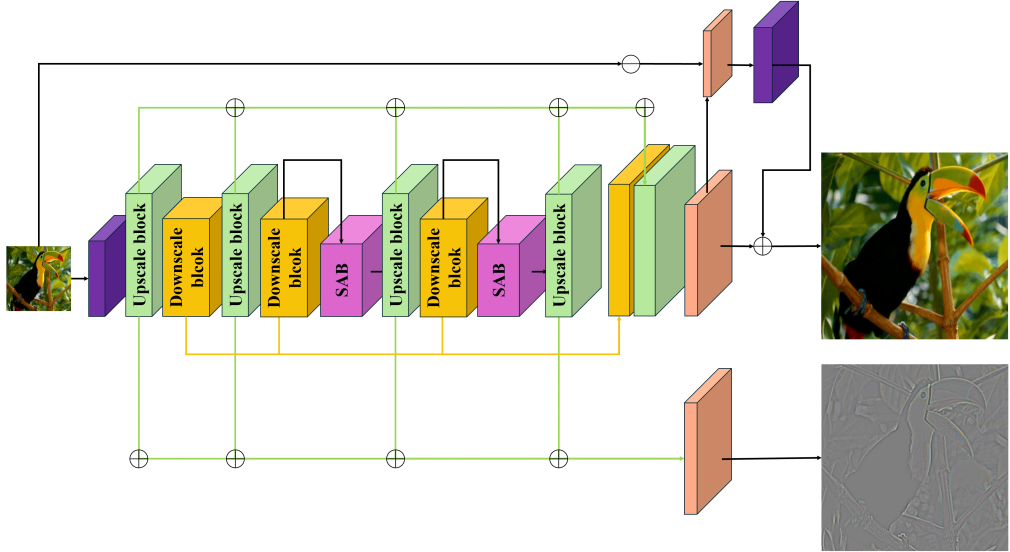

(b) The modified ABPN.

**FIGURE A3** The original ABPN structure and modified version to apply our methods. ((a) reproduced from [21] © 2019 IEEE; (b) adapted from the same source; used with permission.)

The model DRLN is trained for 3000 epochs with a batch size of 16, using a dataset combining DIV2K and Flickr2K. During training, images with pixel values ranging from 0 to 255 are randomly cropped to $48 \times 48$ for LR and $(48 \times s) \times (48 \times s)$ for HR images, where $s$ denotes the upscaling factor. Data augmentation includes random horizontal flips, vertical flips, and 90-degree rotations. The initial learning rate is set to $10^{-4}$ and halved every 200 epochs. When our LP-based detail loss and RUDP are applied to DRLN, the weights used for the weighted sum of losses are identical to those set in LaUD.

For HAN, training is implemented using images from the DIV2K dataset with pixel values in the range of 0 to 255. Since RCAN is used as a pretrained model, only 400 epochs of training are performed with a batch size of 16. The learning rate setup, the cropped LR and HR sizes, and the data augmentation are identical to those used in DRLN. When training with our LP-based detail loss, we set the weight between the SR loss and detail loss to a 1:1 ratio, consistent with the configuration used in LaUD.

**TABLE B1**   The performance variations with different numbers of upscaling steps for RUDP. All models are designed for the 2× super-resolution problem.

| Number of | Set5 | | Set14 | | BSD100 | | Urban100 | |
|---|---|---|---|---|---|---|---|---|
| upscaling | PSNR | SSIM | PSNR | SSIM | PSNR | SSIM | PSNR | SSIM |
| 1 | 38.1402 | 0.9613 | 34.0151 | 0.9209 | 32.3038 | 0.9015 | 32.5285 | 0.9396 |
| 2 | 38.2802 | 0.9619 | 34.4278 | 0.9238 | 32.4483 | 0.9035 | 33.4187 | 0.9476 |
| 3 | 38.4237 | 0.9625 | 34.7677 | 0.9256 | 32.5504 | 0.9045 | 34.0834 | 0.9529 |

**TABLE B2**   The performance across different loss functions. All models are designed for the 2× super-resolution problem.

| Loss | | Set5 | | Set14 | | BSD100 | | Urban100 | |
|---|---|---|---|---|---|---|---|---|---|
| SR | Detail | PSNR | SSIM | PSNR | SSIM | PSNR | SSIM | PSNR | SSIM |
| $l_1$ | | 38.3154 | 0.9620 | 34.6050 | 0.9250 | 32.4888 | 0.9037 | 33.6879 | 0.9497 |
| $l_1$ | $l_2$ | 38.3899 | 0.9627 | 34.5614 | 0.9252 | 32.5058 | 0.9041 | 33.8258 | 0.9514 |
| $l_1$ | $l_1$ | 38.4237 | 0.9625 | 34.7677 | 0.9256 | 32.5504 | 0.9045 | 34.0834 | 0.9529 |

Finally, ABPN differs slightly from the previous two models, as its smallest upscaling factor is 4. The model is trained using the DIV2K and Flickr2K datasets for 5000 epochs with a batch size of 16. HR images are set to a size of $160 \times 160$, while LR images are $40 \times 40$. Only random horizontal and vertical flips are applied as augmentation. The initial learning rate is set to $10^{-4}$, consistent with the previous two models, but it is halved only once after 2500 epochs. Unfortunately, when applying our methods, we are unable to incorporate the weighted sum of losses after RUDP. As a result, only the weight between the SR loss and detail loss is set to a 1:1 ratio.

## C   ADDITIONAL ABLATION STUDIES

In this section, we present additional ablation studies for our two methods: the LP-based detail loss and RUDP. These experiments are conducted using our LaUD. In this study, the test datasets include Set5 [52], Set14 [53], BSD100 [54], and Urban100 [62], as used in the main text.

### C.1   Ablation on Upscaling and Downscaling Repetitions
The number of upscaling steps for RUDP is a hyperparameter. In LaUD, it is fixed at three, a choice determined through extensive experiments, including the ablation study presented in this subsection. Table B1 presents the model's performance according to the number of upscaling steps for RUDP. We evaluate LaUD by varying the upscaling steps to 1, 2, and 3. All models in this experiment are trained once on the ImageNet dataset with our LP-based detail loss.

In summary, increasing the number of upscaling steps consistently results in higher PSNR and SSIM values across all datasets. For PSNR, an improvement of at least 0.1 dB is observed in every case as the number of upscaling increases. Notably, for Urban100, increasing the upscaling steps from 1 to 2 yields a substantial improvement of nearly 1 dB. Although the trained models were not recorded due to data loss, performance improvements become negligible once the number of upscaling steps reaches four or more, and in some cases, the model even performs worse. Moreover, increasing the number of upscaling steps significantly increases resource requirements, including training time and memory usage (details are provided in Section C.3). Based on these experimental observations, we determine the number of upscaling steps that yield fine performance within our resource constraints. Consequently, our LaUD is configured with three upscaling processes.

### C.2   Comparison of Loss Functions
Although we do not directly compare variations of the loss function in the main text, the ablation study (cf. Table 4 of the main text) provides valuable insight into the differences between the standard $l_1$ loss, commonly employed in SR tasks, and our proposed loss function, which combines the standard $l_1$ loss with our LP-based detail loss. As demonstrated by the comparisons between M1 and M2, as well as M3 and M4 in Table 4 of the main text, models utilizing the combined loss function consistently achieve higher performance. For clarity, we restate the results of our LaUD without the LP-based detail loss (model M3) and with the LP-based detail loss (model M4) in the first and third rows of Table B2, respectively.

In addition, we conduct an additional experiment with a slight modification to our combined loss function, as shown in **TABLE** C3   Comparison of model size, memory usage, training time, and inference time with different numbers of upscaling steps in LaUD. "M" indicates million, "MiB" denotes mebibyte, and "ms" denotes millisecond.

| Number of upscaling | Number of Parameters (M) | Memory Usage (MiB) | Training Time (s) | Inference Time (ms) |
|---|---|---|---|---|
| 1 | 8.40 | 868.0 | 15.9587 | 3.686 |
| 2 | 18.57 | 1408.0 | 34.6022 | 7.350 |
| 3 | 29.33 | 1982.0 | 53.5079 | 11.951 |
| 4 | 40.69 | 2574.0 | 73.1315 | 13.746 |

**TABLE** C4   Comparison of model size, memory usage, training time, and inference time of LaUD with other SOTA models. "M" indicates million, "MiB" denotes mebibyte, and "ms" denotes millisecond.

| Model | Number of Parameters (M) | Memory Usage (MiB) | Training Time (s) | Inference Time (ms) |
|---|---|---|---|---|
| EDSR [14] | 40.73 | 1348.0 | 45.0799 | 6.940 |
| D-DBPN [46] | 5.95 | 1350.0 | 27.4748 | 9.934 |
| RCAN [50] | 15.44 | 1714.0 | 174.1959 | 43.491 |
| DRLN [19] | 34.43 | 1952.0 | 63.2655 | 18.058 |
| HAN [22] | 15.92 | 1856.0 | 172.0624 | 44.417 |
| EDT-B [23] | 11.48 | 8070.0 | 383.7992 | 84.582 |
| SwinFIR [25] | 14.35 | 5742.0 | 161.3138 | 66.072 |
| HAT-L [26] | 40.70 | 13146.0 | 331.3637 | 90.863 |
| LaUD | 29.33 | 1982.0 | 53.5079 | 11.951 |

the second row of Table B2. Specifically, this experiment evaluates a model that retains the $l_1$ loss for SR but replaces the $l_1$ loss for detail with the $l_2$ loss. As Table B2 shows, using the $l_1$ loss for the detail part yields higher PSNR and SSIM on every dataset except one: SSIM on Set5 is marginally lower than with $l_2$ loss.

### C.3   Comparison of Model Complexity
Our model, LaUD, is designed without introducing particularly complex techniques beyond the LP-based detail loss and RUDP. However, increasing the number of upscaling steps in RUDP naturally raises the model's complexity. In light of this, we aim to compare the complexity of LaUD for different upscaling steps and relative to existing state-of-the-art (SOTA) models. Table C3 summarizes the model size, memory usage, training time, and inference time of LaUD for different numbers of upscaling steps. Table C4 provides a similar summary comparing LaUD under our configuration with various SOTA models. From the models listed in Table 1 of the main text, we select those requiring no modifications, as their configurations and model construction codes are publicly available. All experiments are conducted under consistent conditions.

Memory usage and training time are measured using an input image with a size of $2 \times 3 \times 64 \times 64$. Generally, larger batch sizes are preferred during training, so initially, a higher batch size was considered for measurement. However, for models such as EDT-B, SwinFIR, and HAT-L, memory requirements exceeded our available resources. As a result, the batch size is standardized to 2 for all models. Inference time is measured using an input image with a size of $1 \times 3 \times 64 \times 64$. For training, the time is measured over 1,000 iterations. Before all measurements for time, 100 warm-up iterations are executed.

As shown in Table C3, the resource requirements of the model consistently increase with the number of upscaling steps (denoted by $K$ in the main text). Specifically, with each increment in $K$ from 1 to 4, the number of parameters increases by approximately 10-11 M, and memory usage grows by around 500-600 MiB. Additionally, the training time increases by roughly 20 s with each increment in $K$. In particular, we draw attention to the comparison of the complexity of LaUD with $K = 4$ (Table C3) against various SOTA models presented in Table C4. Regarding training and inference time, LaUD with $K = 4$ remains efficient enough to compete with other SOTA models, requiring only 73.1315 s and 13.746 ms, respectively. However, LaUD with $K = 4$ exhibits the highest level of parameter count at 40.69 M and memory usage of 2574 MiB, consuming considerably more memory than most other SOTA models, which typically require between 1300 and 1900 MiB, with the exception of EDT-B, SwinFIR, and HAT-L. Considering these resource demands and the previous ablation results on upscaling and downscaling repetitions in Section C.1, we determine that three upscaling steps represent the most suitable configuration. Therefore, LaUD is constructed with $K = 3$ in the main text.

As noted earlier, our LaUD presented in the main text is configured with three upscaling steps. Table C4 contrasts its computational complexity with recent SOTA models; with 29.33 M parameters, LaUD sits roughly in the middle of the group. Although the properties of RUDP make the model slightly heavier, LaUD achieves the highest performance among models excluding the top three (EDT-B, SwinFIR, and HAT-L; see Table 1 of the main text). Moreover, LaUD demonstrates substantial advantages when evaluated against other comparative metrics, as detailed below. Regarding memory usage, the top three models require an exceptionally large amount of memory. In contrast, LaUD occupies only 1982 MiB, which is comparable to smaller models such as RCAN and HAN, enabling it to operate without memory-related concerns. In terms of training time, EDT-B and HAT-L require the longest durations, averaging around 350 s. RCAN, HAN, and SwinFIR follow, requiring approximately half that time. In contrast, LaUD demonstrates significantly faster training at just 53.51 s, highlighting the model's computational efficiency. The overall trend is similar for inference time. Notably, LaUD requires only 11.951 ms, which is comparable to D-DBPN that has a significantly smaller model size.

In summary, the results presented in Table C4 highlight that LaUD is an efficient model. Although the inclusion of RUDP increases the model's complexity, its number of parameters remains reasonable considering the performance achieved. Furthermore, LaUD shows considerable strength in memory usage as well as training and inference times. Consequently, we think that these comparisons indirectly underscore the efficiency of our LP-based detail loss and RUDP.

## D ADDITIONAL IMAGE RESULTS

### D.1 Result Image Examples of LaUD
Figure D4 presents two example image results of LaUD for $2\times$ SR. In both cases, the PSNR values increase progressively with each stage of RUDP. Based on the detail images $D_{SR_k}, k = 1, 2, 3$ generated by LaUD, the final SR image $I_{SR_3}$ exhibits enhanced high-frequency details, such as repeated patterns and clearer object boundaries. Additional examples are available in our GitHub repository.

### D.2 Additional Images for Feature Map Analysis of LaUD
In Figure 3 of the main text, we present a subset of the SR feature maps, highlighting the impact of the LP-based detail loss and RUDP. Figure D5 and Figure D6 below extend this analysis by displaying not just a subset but all 256 channels, using examples from Set5.

The observed trend is consistent with the description provided in the main text. In both cases (a) and (b), more apparent feature maps are generated as the upscaling process is repeated through RUDP. Notably, in case (a), where the LP-based detail loss is applied, more channels are activated during the initial upscaling compared to case (b). Furthermore, when comparing the final SR feature maps between (a) and (b), significantly more features remain prominently active in (a) without fading. This clearly demonstrates the impact of our methods, which help the model extract more diverse high-frequency information and further enhance it to produce a finer SR image.

### D.3 Additional Visual Comparisons of LaUD
Figure D7 shows additional visual comparisons on Urban100. "Image016" depicts a building with a vertical line pattern. In the red-boxed area, although the wall texture is visible in the HR image, none of the models can accurately reconstruct this detail. A closer inspection of the window frame at the bottom reveals differences. D-DBPN generates a blurred result, resulting in the lowest performance. In the right part of the frame, compared to DRLN, SwinFIR, and HAT-L, our model highlights white pixels reflecting light more effectively.

The image, "Image045," features a repeating vertical straight-line pattern. In the region highlighted by the red box, our LaUD ranks third, after HAT-L and DRLN. Overall, our model produces an image with less blur in the middle section compared to D-DBPN and SwinFIR, further highlighting LaUD's ability to preserve sharp, repetitive structures across diverse scenes.

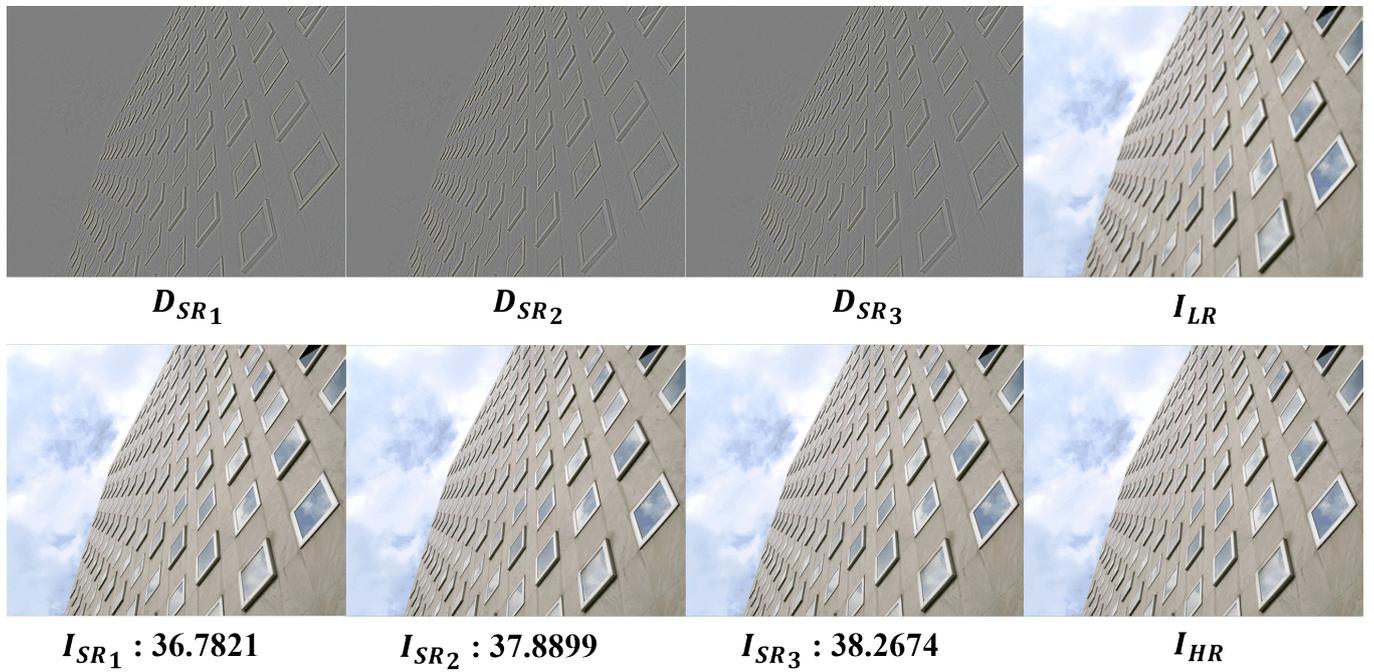

Image010 from Urban100.

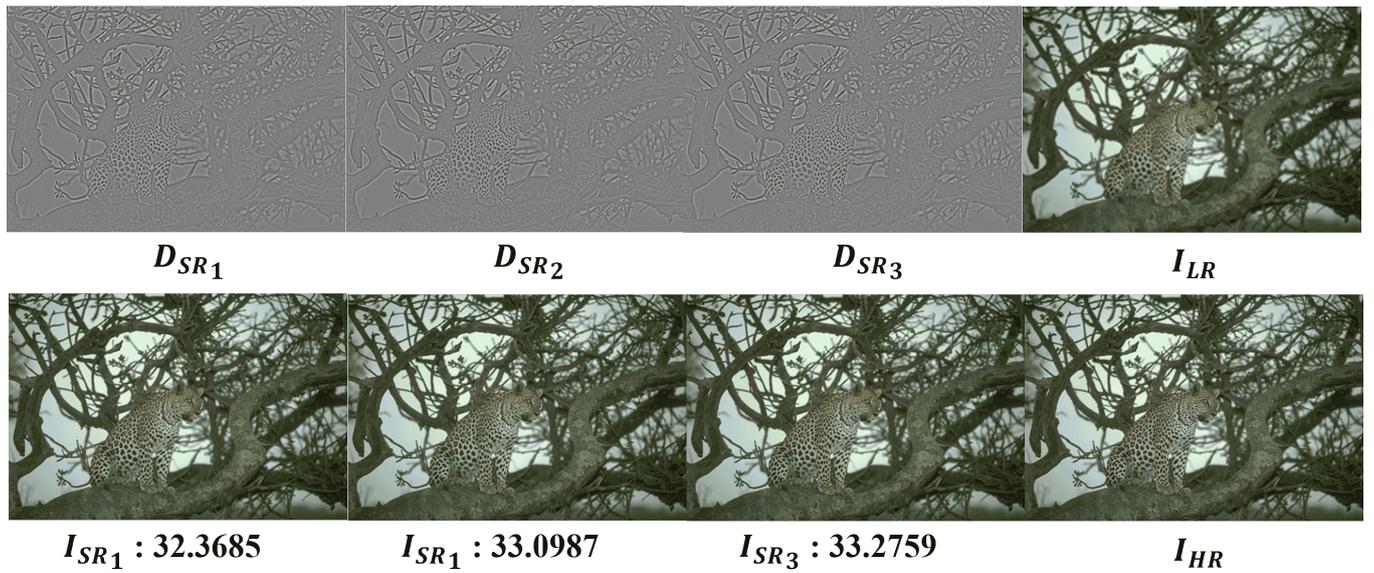

Image016 from BSD100.

**FIGURE D4**  Two example image results of LaUD. Each example is taken from Urban100 and BSD100, respectively. In the first row, there are three detail images of LaUD and the low-resolution image. In the second row, there are SR images of LaUD and the high-resolution image.

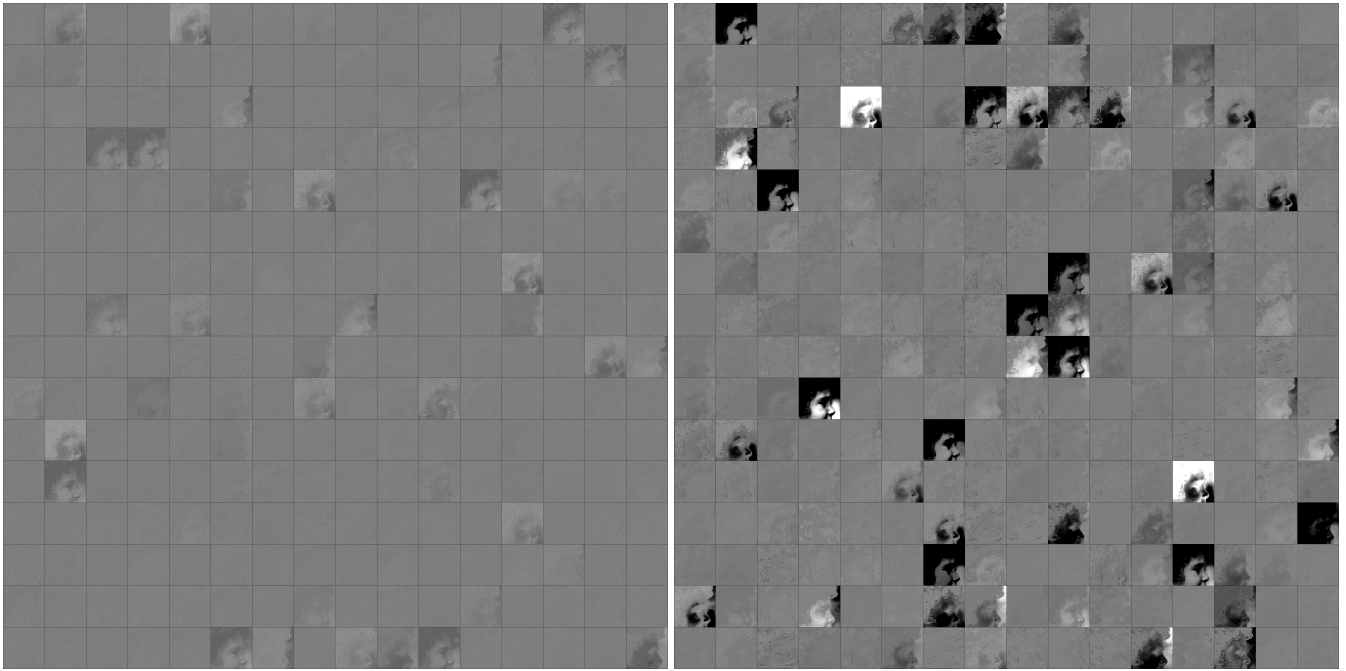

(a) From LaUD.

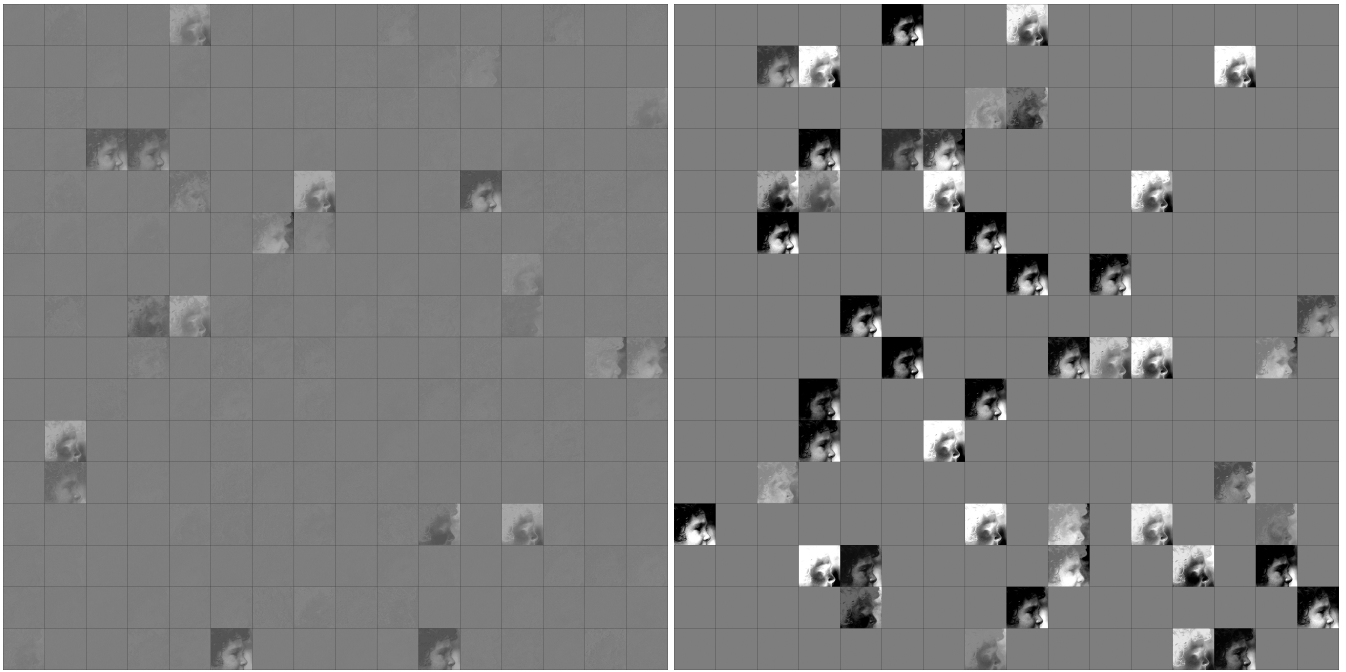

(b) From LaUD without detail loss.

**F I G U R E  D5**  The full SR feature map (for 2×) of the same image, "baby," shown in Figure 3 of the main text: (a) from LaUD, and (b) from LaUD without detail loss. In each row, the left image shows the result after initial upscaling, while the right image shows the result after final upscaling in RUDP.

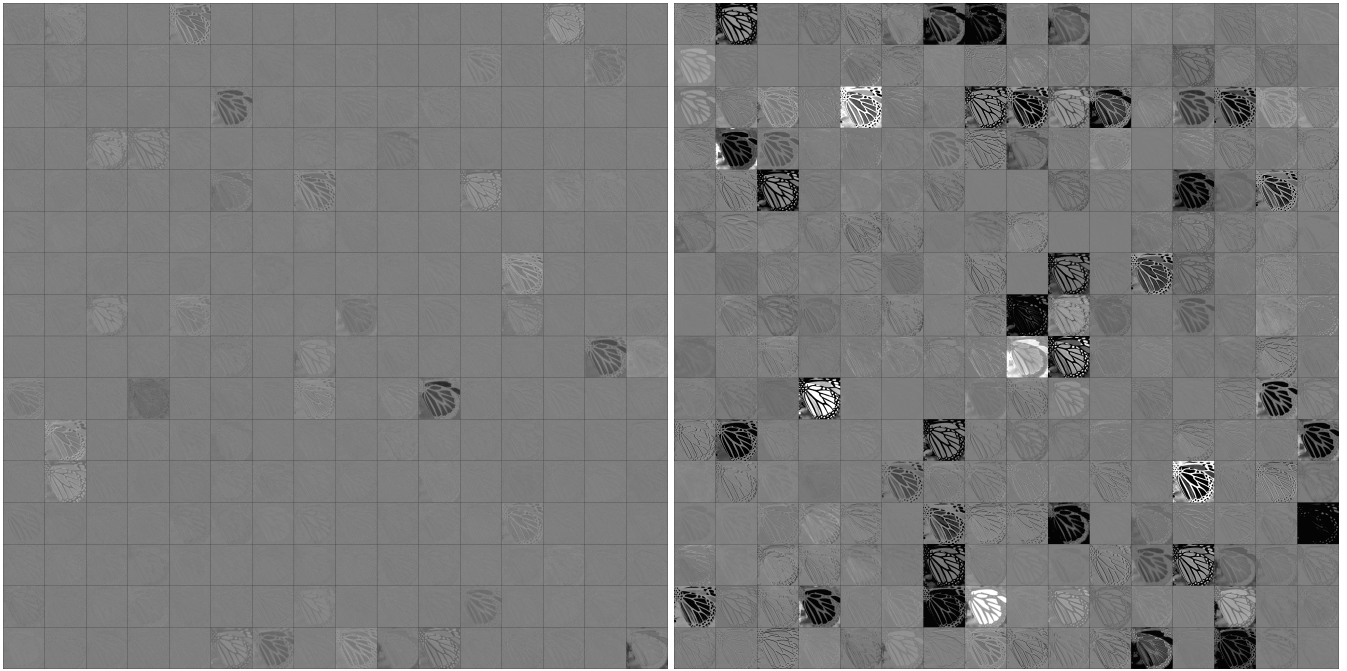

(a) From LaUD.

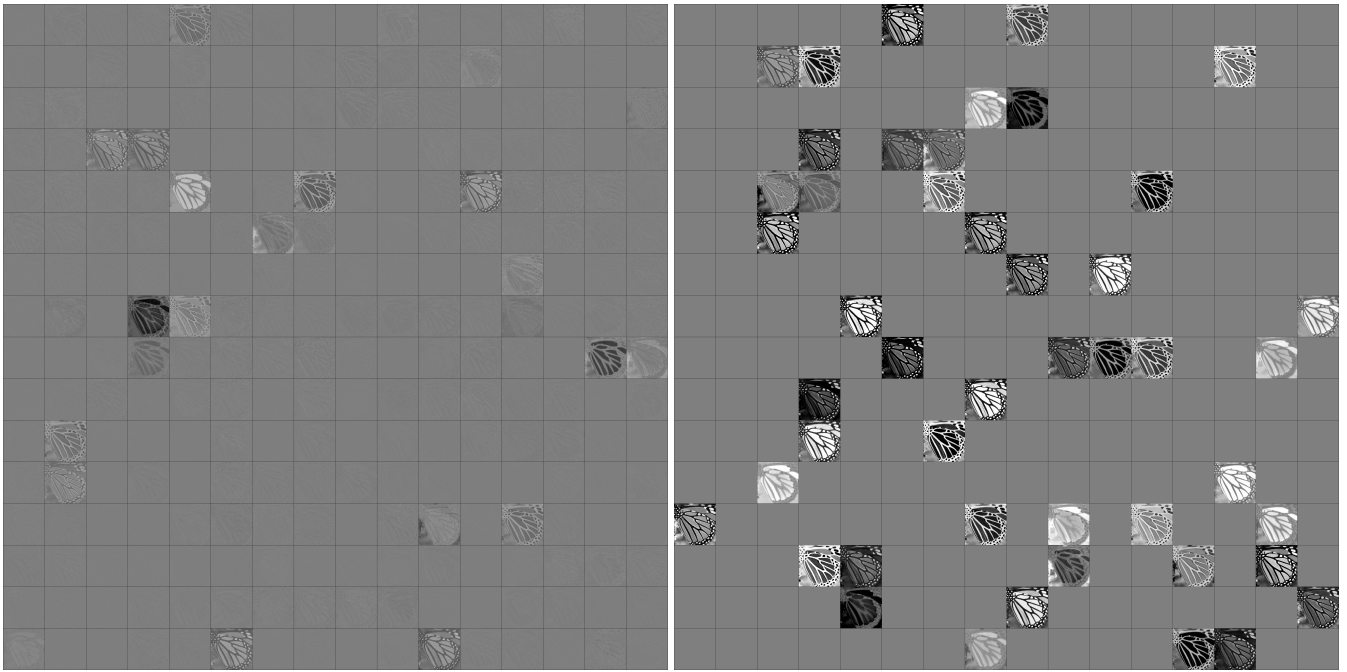

(b) From LaUD without detail loss.

**FIGURE D6**  The full SR feature map (for 2×) of another example, "butterfly," on Set5: (a) from LaUD, and (b) from LaUD without detail loss. In each row, the left image shows the result after initial upscaling, while the right image shows the result after final upscaling in RUDP.

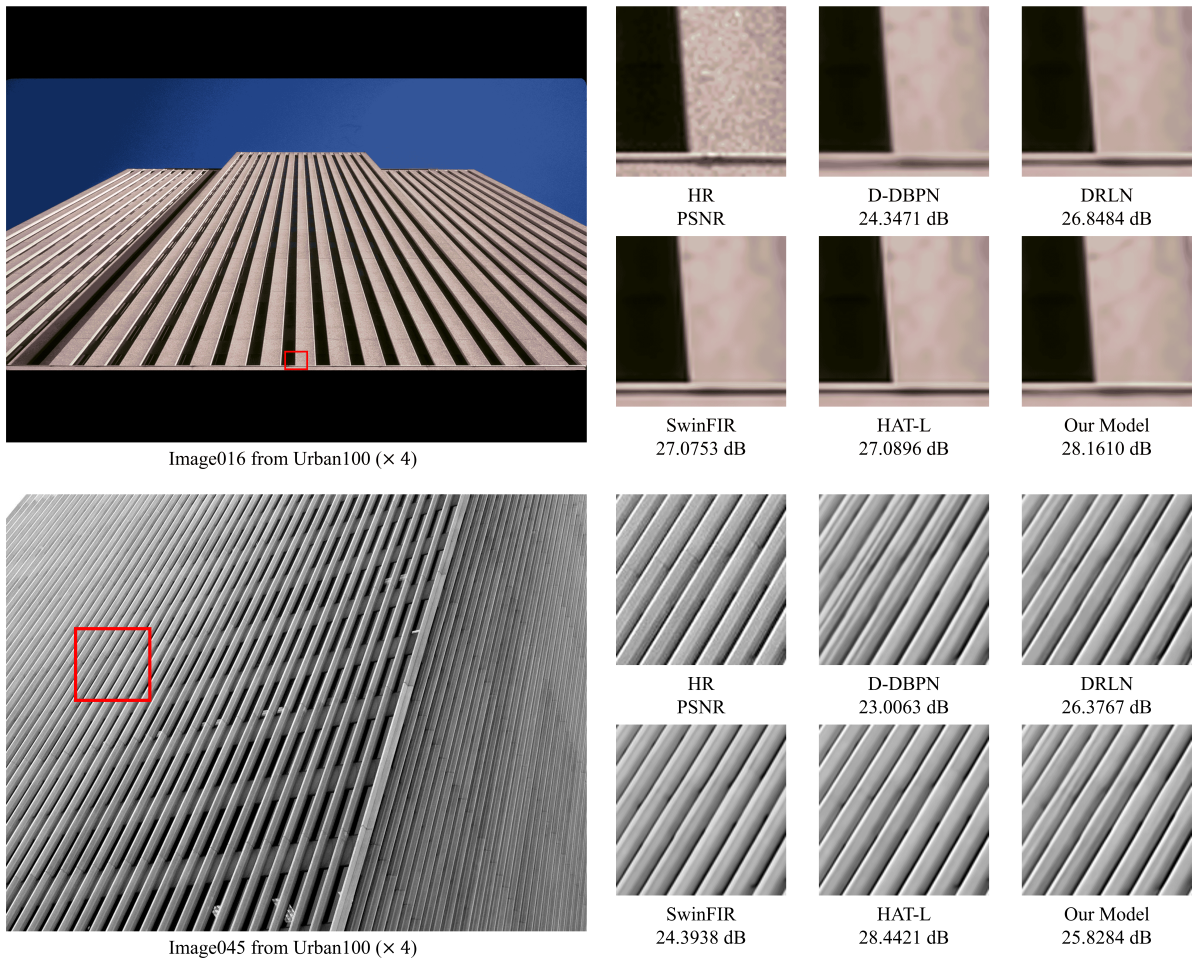

**FIGURE D7** Additional visual comparisons for 4× SR on Urban100. The patches for comparison are marked with red boxes in the original images. The PSNR values below the patches are calculated based on the patches.